\documentclass[journal,comsoc]{IEEEtran}
\usepackage[T1]{fontenc}% optional T1 font encoding

\ifCLASSINFOpdf
\else
\fi
\usepackage{amsmath}

\usepackage[cmintegrals]{newtxmath}
\usepackage{graphicx}
\usepackage{algorithm}
\usepackage{algorithmic}
\usepackage{subfigure}
\usepackage{multirow}
\usepackage{wrapfig}
\usepackage{color}
\usepackage{bbding}
\usepackage[citecolor=red]{hyperref}

\begin{document}

\title{Attentional Graph Convolutional Network for Structure-aware Audio-Visual Scene Classification}

\newcommand{\orcid}[1]{\href{https://orcid.org/#1}{\includegraphics[scale=0.08]{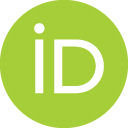}}}

\author{Liguang~Zhou\orcid{0000-0003-0237-1377},~\IEEEmembership{Student Member,~IEEE,}
        Yuhongze~Zhou,~%~\IEEEmembership{Fellow,~OSA,}
        Xiaonan~Qi,~%~\IEEEmembership{Fellow,~OSA,}
        Junjie~Hu, \\%~\IEEEmembership{Fellow,~OSA,}
        Tin Lun~Lam\orcid{0000-0002-6363-1446},~\IEEEmembership{Senior Member,~IEEE,}
        and~Yangsheng~Xu,~\IEEEmembership{Fellow,~IEEE}% <-this % stops a space
\thanks{Manuscript received September 20, 2021; revised XXXXX XX, XXXX. This paper was partially supported by funding AC01202101103 from the Shenzhen Institute of Artificial Intelligence and Robotics for Society. (Corresponding Author: Tin Lun Lam.)}       
\thanks{L. Zhou, J. Hu, T. Lam, and Y. Xu are with Shenzhen Institute of Artificial Intelligence and Robotics for Society, The Chinese University of Hong Kong, Shenzhen, China, (email:hujunjie@cuhk.edu.cn)}
\thanks{L. Zhou, X. Qi, T. Lam, and Y. Xu are with School of Science and Engineering, The Chinese University of Hong Kong, Shenzhen, China (email:liguangzhou@link.cuhk.edu.cn, xiaonanqi@link.cuhk.edu.cn, tllam@cuhk.edu.cn, ysxu@cuhk.edu.cn)}
\thanks{Y. Zhou is with McGill University, Montr\'{e}al, Canada, (email:yuhongze.zhou@mail.mcgill.ca)}}


\markboth{Journal of \LaTeX\ Class Files,~Vol.~14, No.~8, September~2021}%
{Shell \MakeLowercase{\textit{et al.}}: Bare Demo of IEEEtran.cls for IEEE Communications Society Journals}



\maketitle

\begin{abstract}
Audio-Visual scene understanding is a challenging problem due to the unstructured spatial-temporal relations that exist in the audio signals and spatial layouts of different objects and various texture patterns in the visual images. Recently, many studies have focused on abstracting features from convolutional neural networks while the learning of explicit semantically relevant frames of sound signals and visual images has been overlooked. To this end, we present an end-to-end framework, namely attentional graph convolutional network (AGCN), for structure-aware audio-visual scene representation. First, the spectrogram of sound and input image is processed by a backbone network for feature extraction. Then, to build multi-scale hierarchical information of input features, we utilize an attention fusion mechanism to aggregate features from multiple layers of the backbone network. Notably, to well represent the salient regions and contextual information of audio-visual inputs, the salient acoustic graph (SAG) and contextual acoustic graph (CAG), salient visual graph (SVG), and contextual visual graph (CVG) are constructed for the audio-visual scene representation. Finally, the constructed graphs pass through a graph convolutional network for structure-aware audio-visual scene recognition. Extensive experimental results on the audio, visual and audio-visual scene recognition datasets show that promising results have been achieved by the AGCN methods. Visualizing graphs on the spectrograms and images have been presented to show the effectiveness of proposed CAG/SAG and CVG/SVG that could focus on the salient and semantic relevant regions.
\end{abstract}



%\ifCLASSOPTIONpeerreview
%\begin{center} \bfseries EDICS Category: 3-BBND \end{center}
%\fi

\IEEEpeerreviewmaketitle

\section{Introduction}

Audio-Visual scene understanding is an important research direction for human-robot interaction. Via ambient audio and visual information of the environment, the robot is capable of knowing the environment that it is traveling by. For instance, audio-visual scene understanding has been applied in various applications, including robot hearing \cite{ren2016sound}, relative bearing estimation of robots \cite{basiri2016board}, smart homecare surveillance system \cite{Chen2013}, place recognition for mobile robots \cite{zhuang20123}, image quality assessment \cite{yang2022study}, human activity recognition \cite{do2021soham}, elderly care \cite{do2018rish}, automatic construction \cite{akbal2022learning}, robot discovery of the auditory scene \cite{martinson2007robotic}, environmental understanding for self-driving cars \cite{ni2022improved}, and heart sound classification \cite{li2021heart}.

\begin{figure*}[t]
	\begin{center}
		\subfigure[glass breaking]{\includegraphics[width=4cm]{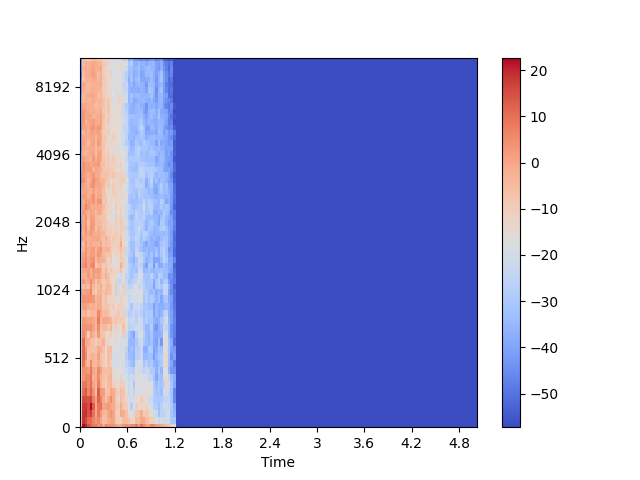}}
		\subfigure[can opening]{\includegraphics[width=4cm]{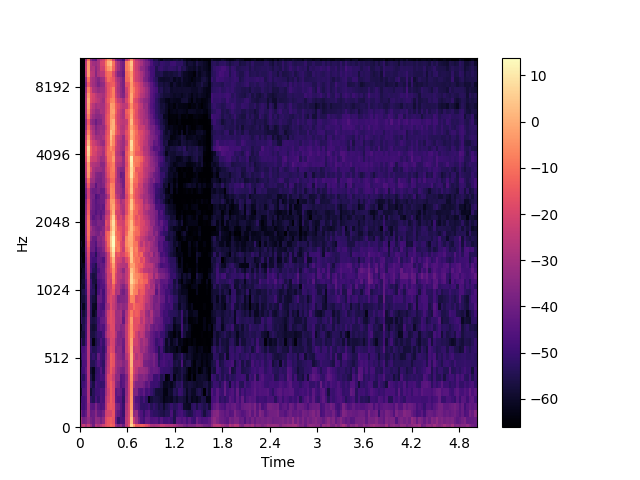}}
		\subfigure[mouse click]{\includegraphics[width=4cm]{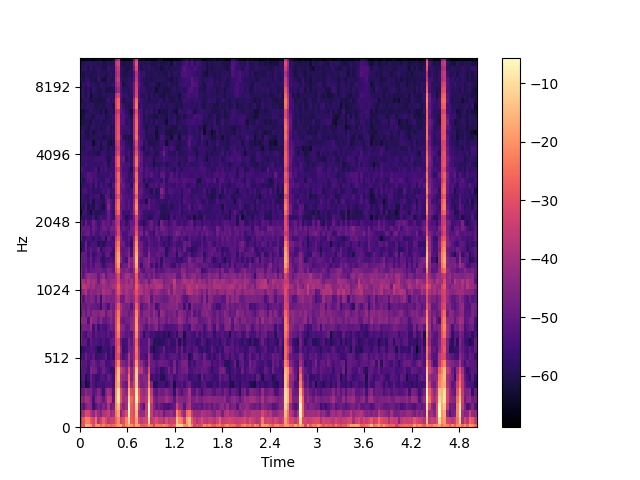}}
		\subfigure[pouring water]{\includegraphics[width=4cm]{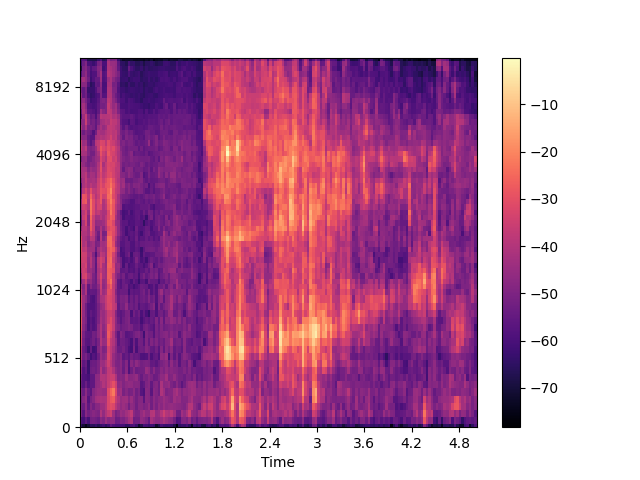}}
		\subfigure[crying baby]{\includegraphics[width=4cm]{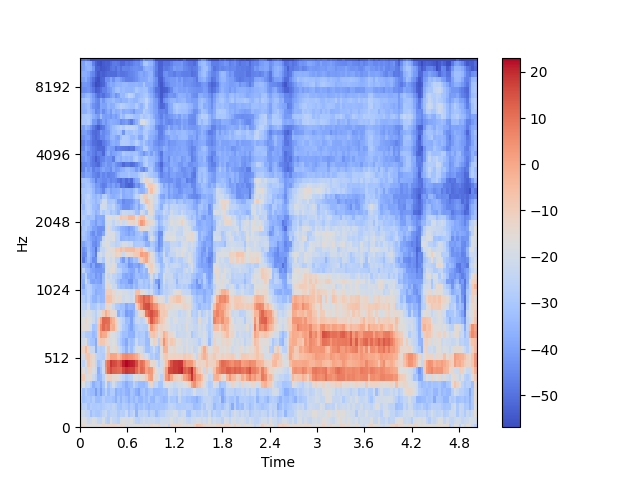}}
		\subfigure[dog]{\includegraphics[width=4cm]{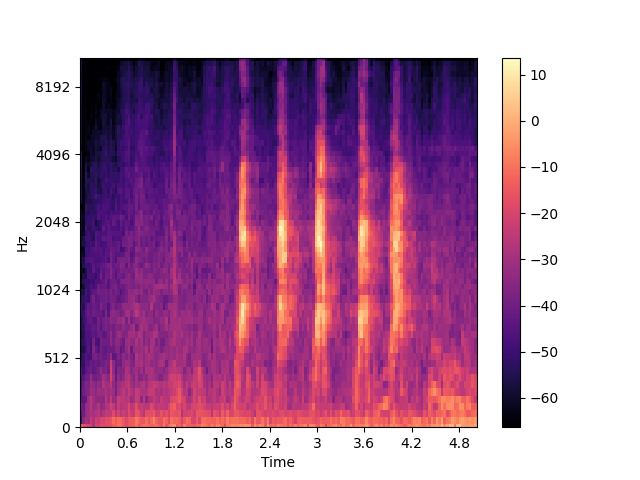}}
		\subfigure[vacuum cleaner]{\includegraphics[width=4cm]{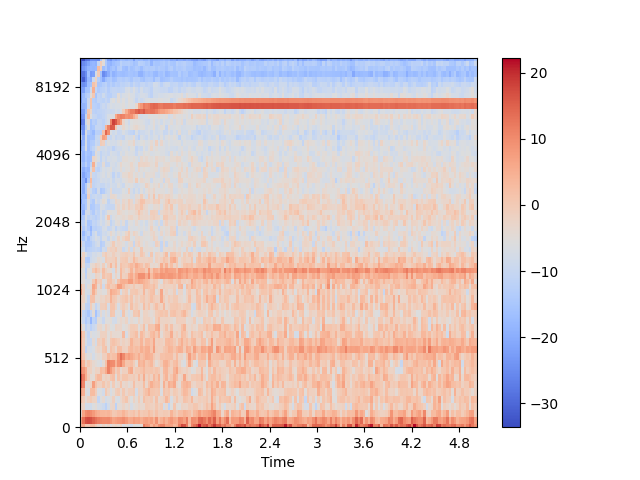}}
		\subfigure[thunderstorm audio]{\includegraphics[width=4cm]{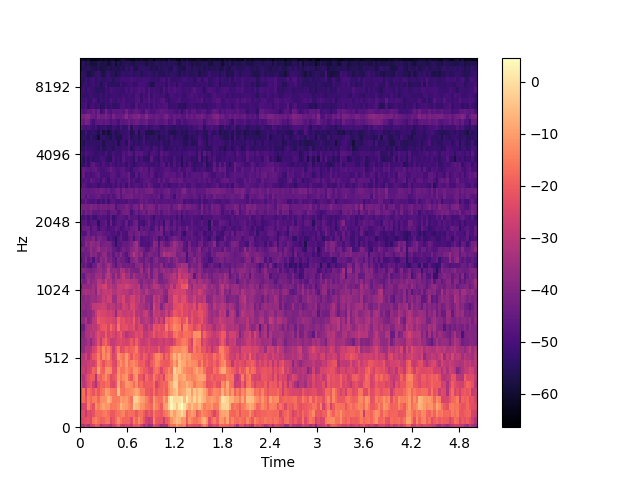}}
	\end{center}\vspace*{-10pt}\caption{Examples of log-Mel spectrograms of various environmental sounds in the ESC-50 dataset.}
	\label{fig:esc50_examples}
\end{figure*}

\begin{figure*}[t]
	\begin{center}
		\subfigure[balcony]{\includegraphics[width=2.2cm]{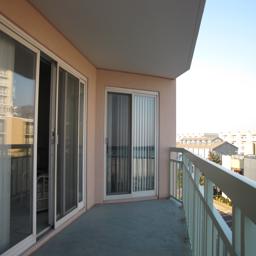}}
		\subfigure[bathroom]{\includegraphics[width=2.2cm]{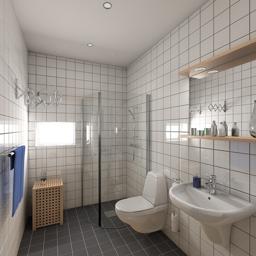}}
		\subfigure[bedroom]{\includegraphics[width=2.2cm]{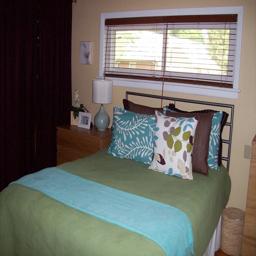}}
		\subfigure[closet]{\includegraphics[width=2.2cm]{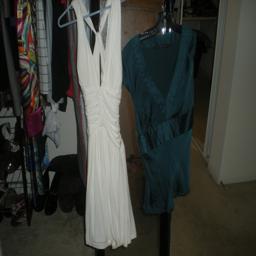}}
		\subfigure[dining room]{\includegraphics[width=2.2cm]{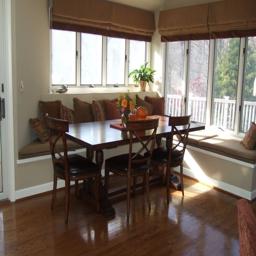}}
		\subfigure[garage]{\includegraphics[width=2.2cm]{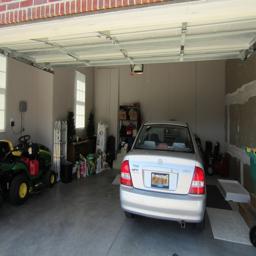}}
		\subfigure[home office]{\includegraphics[width=2.2cm]{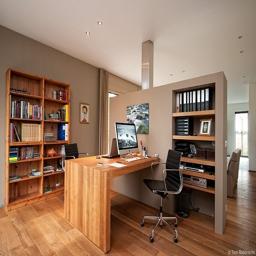}}
		\subfigure[home theater]{\includegraphics[width=2.2cm]{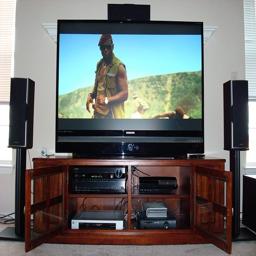}}
		\subfigure[kitchen]{\includegraphics[width=2.2cm]{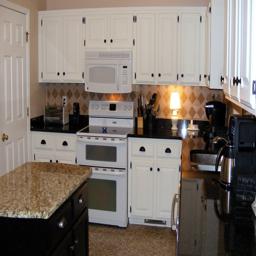}}
		\subfigure[laundromat]{\includegraphics[width=2.2cm]{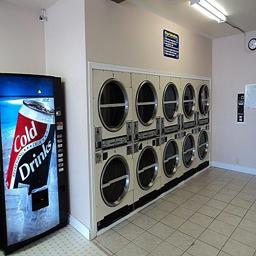}}
		\subfigure[living room]{\includegraphics[width=2.2cm]{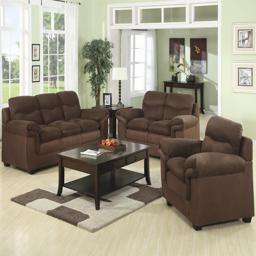}}
		\subfigure[playroom]{\includegraphics[width=2.2cm]{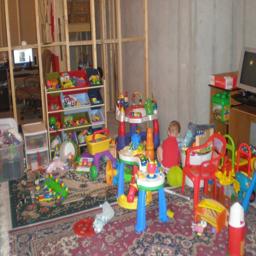}}
		\subfigure[staircase]{\includegraphics[width=2.2cm]{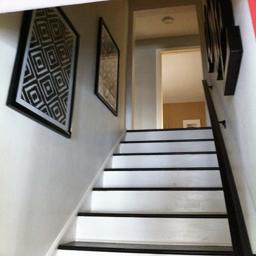}}
		\subfigure[wet bar]{\includegraphics[width=2.2cm]{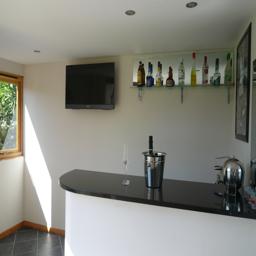}}
	\end{center}\vspace*{-10pt}\caption{Samples of each class of Places365-14 dataset.}
	\label{fig:places365-14}
\end{figure*}

The audio-visual understanding is challenging due to the intrinsic complex nature of audio signals and visual images. For example, the acoustic scene is composed of dynamic and unstructured patterns, which are different and without a regular pattern based on its log-Mel spectrogram as shown in Figure~\ref{fig:esc50_examples}. The visual scene contains various objects spatially distributed over different locations, and different repetitive textures that exist in different scenes as shown in Figure~\ref{fig:places365-14}. To sum up, it is challenging to design an algorithm that effectively captures the texture and characteristics for audio-visual scene representations.

Various efforts have been made to address this challenge for acoustic scene classification (ASC). For example, some works use features like the local binary pattern (LBP) \cite{Toffa2020}, Log-Mel \cite{XinyuWu2009}, wavelet \cite{valero2012gammatone,meintjes2018fundamental}, and Mel-frequency cepstral coefficients (MFCCs) \cite{2009Surveillance}. State-of-the-art solutions on ASC mostly utilize the Log-Mel features \cite{wang2021audio}. To conduct the ASC task, the handcrafted features are first extracted from the audio signal waveform. Then, the trained discriminative models such as support vector machine (SVM) \cite{tripathi2015acoustic} and gaussian mixture models (GMM) \cite{dhanalakshmi2011classification} classify environmental sounds.

Benefiting from the remarkable progress of deep learning \cite{He2016}\cite{wang2015places205}\cite{zhao2022towards}, recent works have exploited the usage of the convolutional neural network (ConvNet) to classify the spectrograms of sounds \cite{piczak2015environmental,wang2021audio}. However, ConvNet only captures local spatial features since convolutional filters focus on the small region of interest (ROI). They neglect the global connection between regions of interest that lies at different locations of visual images or acoustic signals. Further, to capture the salient features that are more relevant to the sound events, attention mechanisms have been proposed \cite{Li2019} \cite{Wang2020}. However, this pooling mechanism only considers the strength of the sound signal but neglects the spatial connection of different sound spectrogram patches.

For visual scene classification (VSC), there has been a longer history and many approaches been proposed, such as object-based methods \cite{espinace2010indoor,li2010objects,zhou2021borm,miao2021object}, bag-of-visual-words \cite{yang2007evaluating}, global image features \cite{meng2012building}, color descriptors \cite{van2009evaluating}, VGGNet \cite{wang2015places205}, ConvNet \cite{Zhou2014}. Until recently, graph convolutional network (GCNs) has been proposed to improve scene recognition \cite{yuan2019acm,khan2019graph,zeng2020amorphous}. These methods have shown the merits of GCNs over ConvNets. However, they only model the most distinctive features or deep semantic/object features without considering the contextual information of backgrounds, which leads to partial information loss.

\begin{figure*}[t]
	\begin{center}
		\includegraphics[width=\linewidth]{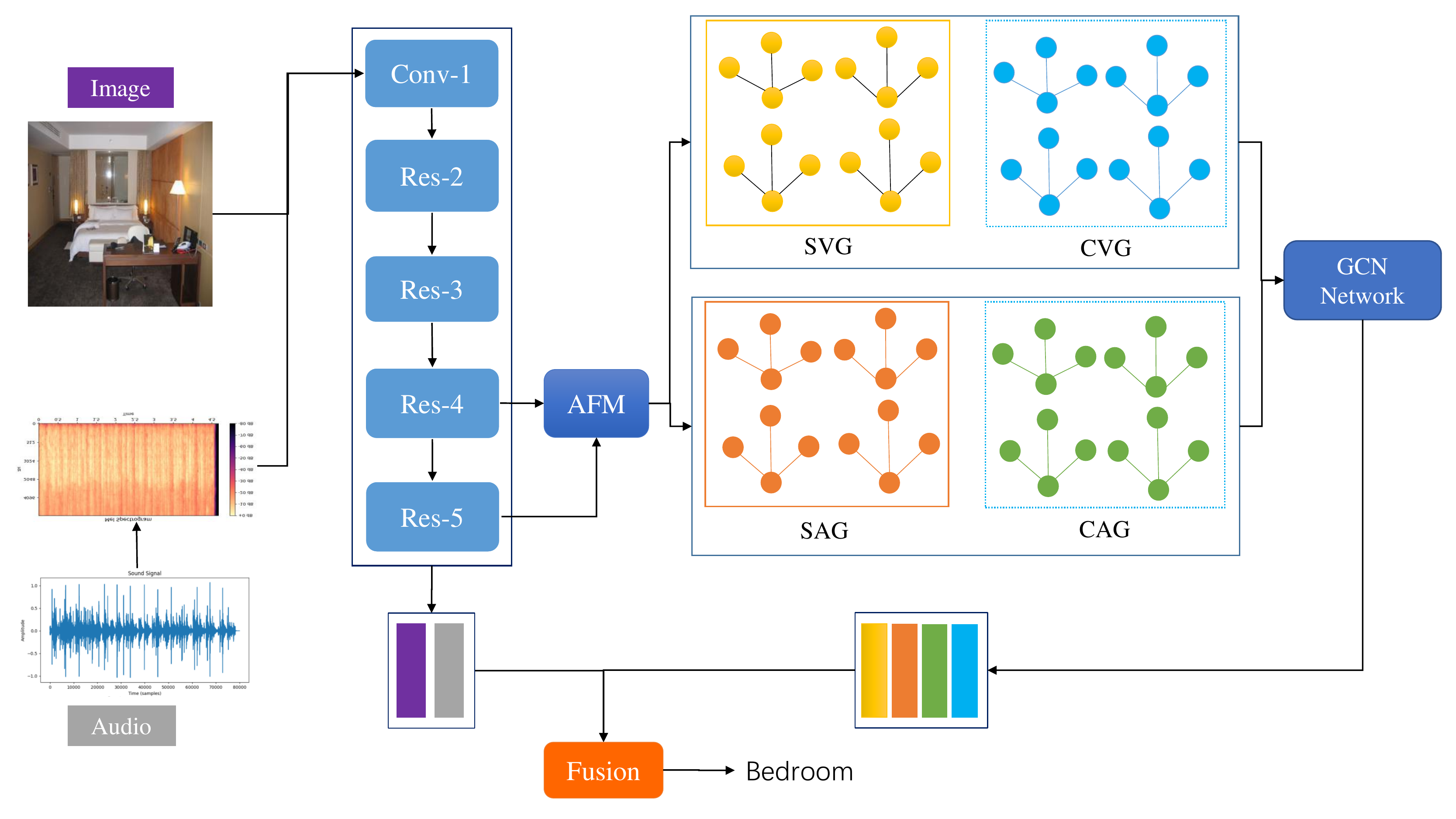}
	\end{center}\vspace*{-10pt}\caption{The proposed audio-visual scene recognition framework. First, the input images and sound spectrograms are processed by a backbone network and intermediate feature maps are obtained. These features are fused by AFM module to expand its feature maps and receptive fields. Then, we constructs the SAG (in orange), CAG (in green), SVG (in yellow), and CVG (in blue) for audio-visual scene representation with a graph convolutional network. Hence, the refined features are obtained. Last, the refined graph features and backbone features are fused for audio-visual scene recognition.}
	\label{fig:method}
\end{figure*}

Different from past works, we present an AGCN for the audio-visual scene classification (AVSC), as shown in Figure~\ref{fig:method}. To represent the audio-visual scene in a structure-aware manner, we leverage GCNs, which could be a suitable but less exploited choice. There are three steps to achieving this. First, to extract feature representation of the audio-visual scene, we use ResNet as a backbone network. Then, to augment the obtained features with multi-scale information, we aggregate the feature maps of ResNet at multiple layers with an attention fusion module \cite{song2021attanet}. Last, to achieve structure-aware audio-visual scene understanding, we construct graphs that can focus on the regions of interest and pay attention to the context information. Specifically, we construct the SAG/SVG that takes the most distinctive nodes selected from the pyramid feature map and the CAG/CVG that takes the average nodes selected from the pyramid feature map. The SAG/SVG is regarded as the distinctive region of interest (ROIs) of audio-visual signals while the CAG/CVG is viewed as contextual information of audio-visual signals. To refine the graph features, these obtained graphs are processed by the GCNs. Finally, the classification results are estimated through the fully connected layer. 

To demonstrate the effectiveness of the proposed method, the experiments are conducted on ASC tasks including ESC-10 and ESC-50 datasets, VSC tasks including Places365-7, Places365-14, and SUN-RGBD datasets, and AVSC task with DCASE2021 datasets, which proves the proposed method reaches comparable performance on these tasks. To further show the difference between the proposed method and the baseline method, the selected CAG/SAG and CVG/SVG are visualized among those audio spectrograms and visual images, which indicates both the semantically relevant regions and contextual regions are noticed during the learning process.

To summarize, our main contribution to this work can be summarized in four folds:
\begin{itemize}
	\item  We propose a simple and effective attentional graph convolutional neural network (AGCN) that achieves structure-aware audio-visual scene classification. 
	
	\item We construct and combine the salient acoustic graph (SAG), contextual acoustic graph (CAG), salient visual graph (SVG), and contextual visual graph (CVG)  to represent the discriminative regions and background contextual regions of audio-visual scene inputs.
	
	\item Extensive experimental results of our method on the ASC, VSC, and AVSC datasets show that our method has achieved comparable performance.
	
	\item Visualization results show the proposed AGCN can focus on both the semantically relevant regions and contextual regions.
	
\end{itemize}

The rest of the paper is organized as follows. The related works are presented in Section~\ref{sec:related_works}. The AGCN network is described in Section~\ref{sec:method}. The experimental results and discussions are shown in Section~\ref{sec:exp}. The conclusion about our research comes in Section~\ref{sec:conclusion}.

\section{Related Works}
\label{sec:related_works}

\subsection{Audio-Visual Scene Classification} 
AVSC is to establish the feature representation of audio-visual input signals and classify it. Hu et al. \cite{hu2020cross} create the ADVANCE for aerial scene recognition with audio-visual scene data. Wang et al. \cite{wang2021curated} introduce a dataset of urban scenes for audio-visual scene analysis.
The AVSC task consists of two modules including the audio module and visual module. For the audio module, there are various networks designed to tackle ASC tasks. In particular, with the rapid development of deep learning in image classification, numerous works have exploited the usage of ConvNet to classify the spectrograms of sounds\cite{piczak2015environmental}. To capture the salient features that are more relevant to the sound events, attention mechanisms have been introduced \cite{Li2019} \cite{Wang2020}. Temporal attention is proposed to capture the temporal structure of sound \cite{Li2019}. A sparse key-point encoding and efficient multispike learning framework are proposed for this problem \cite{yu2020robust}. Parallel temporal-spectral attention is proposed to enhance the representation of both the temporal and spectral features by capturing the importance of different time frames and frequency bands \cite{Wang2020}. These attention mechanisms enhance the feature representations in the channel and spatial level but lack preserving the multi-level feature representation. To address this issue, Zhou et al. \cite{zhou2022feature} shows a feature pyramid attention structure is effective for acoustic scene recognition. However, it neglect the relative position of sound features. To this end, we utilize a latent feature fusion attention module to incorporate the multi-level feature for multi-scale feature extraction. Besides, the GCNs have been utilized for preserving the spatial relation of distinctive and average features of sound patches.

Even though extensive research has been done to improve the performance of ASC, the construction of salient acoustic graphs and contextual acoustic graphs based on graph neural networks is missing from the literature. This is important because the acoustic time-frequency features contain both local and global informative patches for ASC while the convolutional neural network based methods are incompetent of effectively associating these features.

VSC is the task of recognizing the scene with visual inputs.
For the VSC task, there are additional semantic or depth information being incorporated as an additional branch for VSC. For instance, the semantic aided scene recognition become popular in research area. Ni et al. \cite{ni2022improved} propose an improved faster RCNN to extract representative objects as local features for scene recognition. Zhou  \cite{zhou2021borm}  et al. present a Bayesian object pair modeling method to enhance the scene representation ability of network with additional semantic information. Miao  \cite{miao2021object}  et al. develop a branch of scene embedding network  to improve the scene representation. These methods has achieved comparable performance over single branch methods in some extend. However, the semantic branch requires cost expensive computations which limits the applications of these algorithms. To model scene representation with GCNs, an adaptive cross-modal GCNs has been proposed \cite{yuan2019acm}, where the most distinctive features are selected for visual scene representation. Besides, the semantic region masks are selected for graph modeling with GCNs to model the scene representation \cite{zeng2020amorphous}. Segmented regions are utilized to model the remote scene representation in \cite{khan2019graph}. These methods have shown the merits of GCNs over ConvNets on scene representation. However, the background contextual information, which is important for scene representation as well, is abandoned, leading to information loss for scene representation.

To unite the audio and visual modules, we present a novel AGCN method based on the graph convolutional network. The proposed method first augments the feature representation of the backbone network by utilizing an attention-based attention fusion module \cite{song2021attanet}. Especially, salient graphs and contextual graphs are proposed to focus on the most distinctive regions and contextual regions of both audio and visual features.

\subsection{Graph Convolutional Networks}
Graph convolutional network is designed for non-Euclidean and graph-structured data, such as text documents, bioinformatic data, and scene signal \cite{henaff2015deep}\cite{kipf2016semi}. GCNs have been successfully applied in various tasks, including scene recognition \cite{khan2019graph,yuan2019acm}, action recognition \cite{yan2018spatial}, image recognition \cite{chen2019multi}, and emotion recognition \cite{nie2020c}. However, the application of GCNs in audio signal analysis remains underexplored \cite{sun2020ontology,wang2020modeling}. There are two major types of data processing methods in graph convolutional networks, spatial \cite{yan2018spatial} and spectral \cite{yuan2019acm}. For the spatial graph, the convolutional filters are directly applied to the nodes and neighbors, which is efficient but ignores locality information. For the spectral graph, the convolutional filters are performed on the spectral domain, where the data nodes are processed with the Laplacian matrix and convolutional filters. In this paper, we use the spectral graph for graph feature refining. 

\section{Methodology}
\label{sec:method}

As shown in Figure~\ref{fig:method}, our proposed AGCN network mainly contains four steps. The first step is the backbone network for feature extraction where we use ResNet50 as the backbone network. The second step is the graph feature construction, where the attention fusion module (AFM) that refines and aggregates the different levels of feature maps from the backbone network is utilized \cite{song2021attanet}. The third step is graph construction and processing, where we construct salient and contextual graphs for audio-visual feature representation, including the SAG/CAG and the SVG/CVG. Specifically, we utilize the most distinctive and average nodes from the feature maps as a mimic of salient regions and contextual regions for feature abstracting. The detailed graph construction procedure is presented in Algorithm~\ref{alg:K}. Sequentially, these constructed graphs are fed into a spectral graph convolutional network for feature extraction. Last, with a fully connected layer attached, the classification results are obtained.

\begin{figure}[t]
	\begin{center}
		\subfigure{\includegraphics[width=8cm]{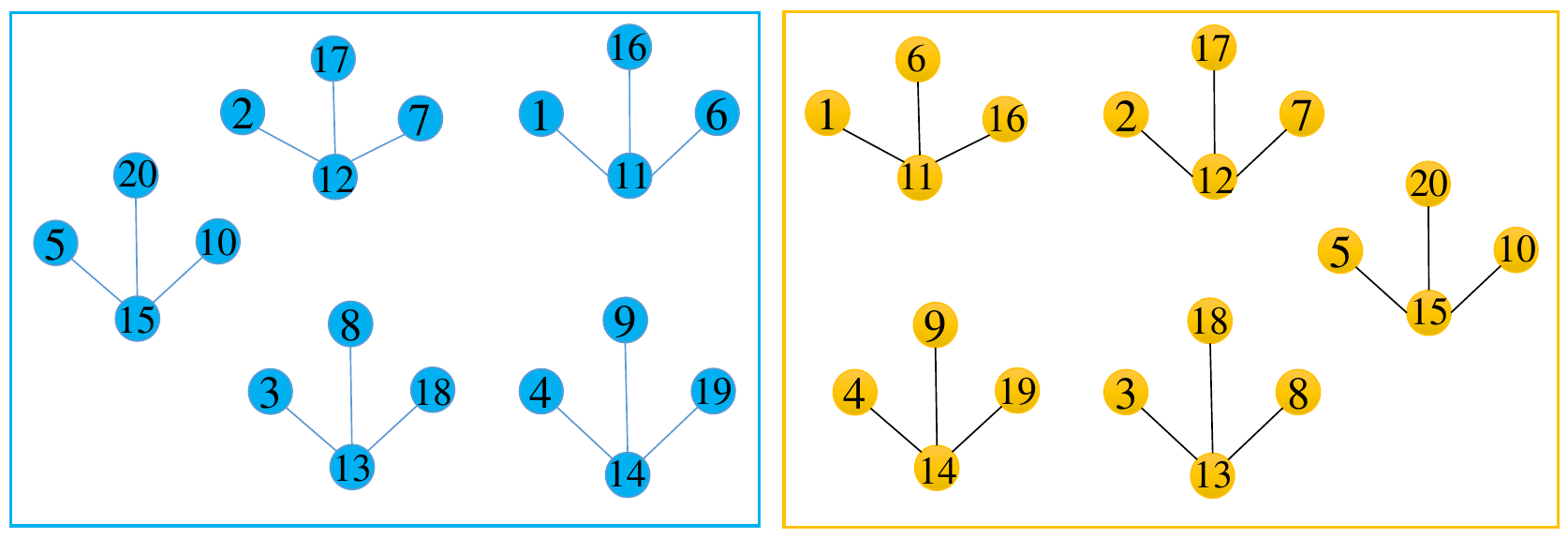}}
		\subfigure{\includegraphics[width=3cm]{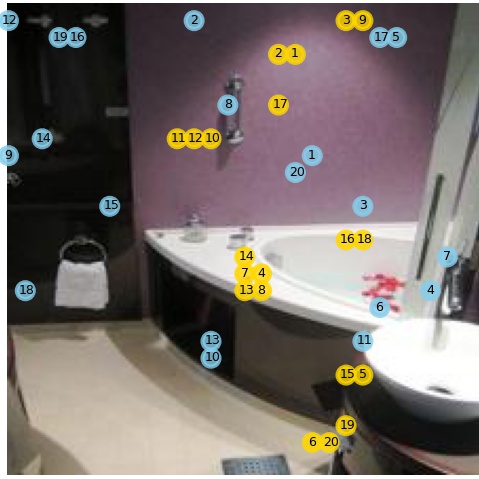}}
		\subfigure{\includegraphics[width=4.2cm]{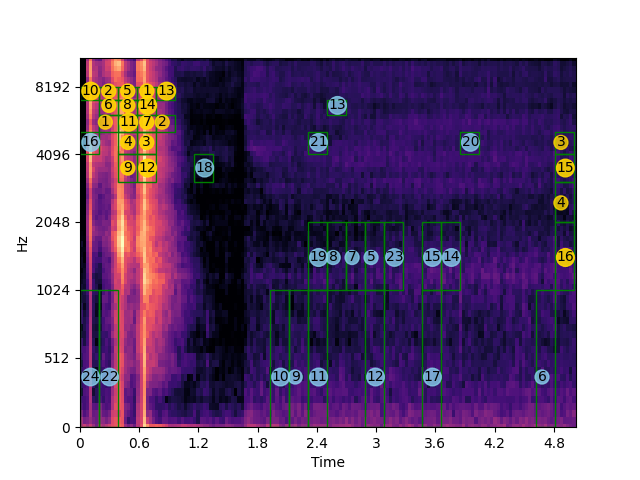}}
	\end{center}\vspace*{-10pt}\caption{In the first row, the left column shows the CAG/CVG in blue, and the right column shows the SAG/SVG in yellow. In the second row, the visualization of CVG and SVG of bedroom is shown in the left column, while the visualization of CAG and SAG of can opening acoustic spectrogram are shown in right column. }
	\label{fig:scene_graph_demo}
\end{figure}

\subsection{Graph Feature Construction}

Consider the acoustic signals as $x_a$. The weights of the ConvNet backbone network and AFM module are represented as $W_{convs}$ and $W_{afm}$ respectively. The input feature $x_a$ will be processed by these two modules consecutively.

\begin{equation}
	F_{m_4}, F_{m_5} = W_{convs}(x_a)
\end{equation}

\begin{equation}
	F_{ffr} = W_{afm}(F_{m_4}, F_{m_5})
\end{equation}

where $F_{ffr}$ is the fused feature representation extracted by the ResNet50 backbone and AFM network.

To select nodes from refined feature maps for graph construction, we first sum the fused feature maps $F_{ffr}$ along the channel dimension as:

\begin{equation}
	F_{rei} = \sum_{i=1}^{C}F_{ffr}(N, i, H, W)
\end{equation}

The tensor shape of $F_{ffr}$ is $(N, C, H, W)$, where $N$ denotes batch size, $C$ represents the number of channels, $H$ is the height of $F_{ffr}$, and $W$ is the width of $F_{afm}$. We propose to select $K$ most distinctive and $K$ average feature vectors to form salient scene graph and contextual scene graph for the graph nodes $V$.

Then the reshaped intensity map $F_{rei}$ is reshaped to $(N, H * W)$, where each pixel shows the intensity of features. As acoustic signal contains natural spatial-temporal relationships, these relationships are essential to ASC. Similarly, visual images contain the intrinsic spatial layouts of various objects and textures, which is important for visual scene representation. To preserve the spatial information, we constructed graphs for audio-visual scene representation with geometric relations. 

\subsection{Audio-Visual Graph Construction}
The unstructured spatial-temporal relationships of acoustic spectrograms in acoustic scenes are difficult to learn. The spatial layouts of various objects in visual scenes are hard to capture. To address these issues, we decompose the audio-visual signals into two novel graphs, SAG/SVG to represent the most distinctive and semantic relevant regions of audio spectrograms and visual images, and CAG/CVG to represent the background contextual information of spectrograms and visual images. Since we only select the distinctive regions and contextual regions of features, this modeling method can skip most silent and semantically irrelevant information of audio-visual scenes. 

Let's take the composition of SAG/CAG as an example. The SAG/CAG can be represented as $G_{sag} = (V, E, A)$ and $G_{cag} = (V, E, A)$. $V$ represents the representative features selected from $F_{rei}$. $E$ denotes the edges of the graph. $A$ represents the adjacency matrix of the graph, which preserves the relative position of the scene graphs.

\subsubsection{Nodes Selection}

The detailed construction of scene graphs are illustrated in the Algorithm~\ref{alg:K}. K denotes the nodes number of graph. 

\begin{algorithm}[!ht]
\begin{center}
	\caption{\label{alg:K} Algorithm for constructing SAG/CAG, and SVG/CVG.}%title
\begin{flushleft}
	\hspace*{\algorithmicindent} \textbf{Input:} The reshaped intensity map $F_{rei}$; \\
    \hspace*{\algorithmicindent} \textbf{Output:} The indexes of $K$ selected feature vectors $ind_{sel}$;
\end{flushleft}

	\begin{algorithmic}[1]%
		\STATE Sort N reshaped intensity maps $F_{rei}$ with descending order, and select the top $K$ and medium $K$ indexes to $ind_{F}$;
		\STATE $m_{left} = \frac{W\times H}{2} - \frac{K}{2}$;
		\STATE $m_{right} = \frac{W\times H}{2} + \frac{K}{2}$;
		\FOR{$i = 0 \rightarrow N$}
		\STATE $index = sort(F_{rei}(i))$;
		\STATE $ind_{sag}(i) = index(1:K) + index(m_{left}:m_{right})$;
		\ENDFOR
		\STATE Sort the $N$ indexes $ind_{F}$ with ascending order to keep the features with original spatial order;
		\FOR{$i = 0 \rightarrow N$}
		\STATE $ind_{sel}(i) = sort(ind_{F}(i))$;
		\ENDFOR
		\RETURN $ind_{sel}$;
	\end{algorithmic}
\end{center}
\end{algorithm}

The nodes of scene graphs $V = \{ v_{i}|i=1,…,K \}$ are associated with the selected features from $2K$ different locations of $F_{ffr}$, with K locations for SAG and the other K locations for CAG. The nodes selection method is shown in Algorithm~\ref{alg:K}. 
The graph nodes assignment can be formulated as $V_{sag} = \{ v_{i} = F_{ffr}(N, C, i)|i = ind_{sel}(1),…,ind_{sel}(K) \}$ and $V_{cag} = \{ v_{i} = F_{ffr}(N, C, i)|i = ind_{sel}(K+1),…,ind_{sel}(2K) \}$, where $F_{ffr}$ is reshaped to $(N, C, H*W)$, and the shape of graph nodes $V$ is $(N, K, C)$.

\subsubsection{Subgraph Construction}
The total number of nodes of SAG/CAG is K. The SAG/CAG is made of N/4 subgraphs (sub-textures), that is, each subgraph is composed of 4 nodes. The subgraph is automatically constructed followed by the subgraph set $G_{sub}$ and the center set of subgraph $G_{csub}$:

\begin{equation}
	\begin{aligned}
		\begin{split}
			 G_{sub}  = & \{ i, i+\frac{1}{4} K, i+\frac{2}{4}K, i+\frac{3}{4}K \mid  \\       
			& i \in [1,K/4], i \in \mathbb{Z}, K \in \mathbb{Z}, K/4==0 \}  \\
			 G_{csub} = & \{ i + \frac{2}{4}K \mid   \\ 
			& i \in [1,K/4], i \in \mathbb{Z}, K \in \mathbb{Z}, K/4==0 \} \\
		\end{split}
	\end{aligned}
\end{equation}

Figure~\ref{fig:scene_graph_demo} shows the case when the node number of the graph is 20 and there are 5 subgraphs with the center of each subgraph assigned 11, 12, 13, 14, and 15. In the first row, 40 distinct nodes are selected from the feature map to construct the SAG and CAG, where 20 most distinctive nodes construct a SAG/SVG in gold and 20 average nodes construct a CAG/CVG in blue. 
In the second row, the proposed graphs are visualized on a "\textbf{bedroom}" image and a "\textbf{can opening}" acoustic signal. It is obvious that the nodes of SVG/SAG mostly locate in the most distinctive area of visual image and sound signals, which indicates the importance of this region. Besides, the nodes of the CVG/CAG are scatted around the whole image and spectrograms to preserve the small details of the contextual background of the visual image and sound event. The combination of these two graphs can well represent the visual image and acoustic signals across the entire spectrograms of both representative objects and background contextual information.

\subsubsection{Geometric Relation}
Since the spatial-temporal relationship of acoustic signals and spatial layouts of visual images are essential for AVSC, to utilize this relationship, we construct the adjacency matrix A according to the geometric relations $\mathbf{E_{ij}}$ among adjacent nodes. 

\begin{equation}
	A_{ij} = 
	\begin{cases}
		\left|x_i-x_j\right|+\left|y_i-y_j\right|, E_{ij}\neq 0 \\ 
		0, E_{ij} = 0
	\end{cases}
\end{equation}

where $(x_i, y_i)$ and $(x_j,y_j)$ represent the position of i-th node and j-th node. $E_{ij} \neq 0$ means that there is a connection between i-th node and j-th node.

\subsection{Spectral Graph Convolution}

Given the acoustic graphs $G(V, E, A)$, the Laplacian matrix L is obtained by L=D-A, where D is degree matrix of graph $G$, and A is the adjacency matrix that records the geometric relations of nodes. To normalize the Laplacian matrix, we use the symmetric normalized Laplacian matrix, 

\begin{equation}
	L^{sym}=D^{-\frac{1}{2}}LD^{-\frac{1}{2}}=I_{k}-D^{-\frac{1}{2}}AD^{-\frac{1}{2}}
\end{equation}

where $I_{K}$ is the identity matrix. With the added self-connections, $\hat A$ is denoted as $\hat A=A+I$. Then, the GCNs follows the same setting as \cite{kipf2016semi}: 
\begin{equation}
	Y=\left(D+I_{k}\right)^{-\frac{1}{2}}\hat A \left(D+I_{k}\right)^{-\frac{1}{2}} X \Theta
\end{equation}

\begin{equation}
	Y=L_{norm} X \Theta_{j}
\end{equation}

Each $\Theta_{j}$ is implemented with the vanilla convolution layer. To reduce the dimension of acoustic graphs, we employ a 1x1 convolution layer and then ReLU is added as the activation function.

\section{Experiments and Discussions}
\label{sec:exp}

In this section, we examine the performance of AGCN on VSC, ASC, and AVSC tasks. Then, the visualization of the proposed SAG/CAG and SVG/CVG are analyzed. These detailed evaluations with extensive results are shown to prove the effectiveness of the proposed AGCN.

\subsection{Visual Scene Recognition}

\subsubsection{Datasets} 
We use the Places365-7 and Places365-14 datasets for the evaluation of the proposed AGCN. The dataset split is the same as BORM \cite{zhou2021borm}. For the Places365-7 dataset, there are 7 classes, 35000 images as the training set, and 701 images as the test set. For the Places365-14 dataset, there are 14 classes, and 75000 images for the training set,  1500 images for the test set. The SUN RGB-D is also utilized. Following the same settings as in \cite{pereira2021deep}. This dataset is composed of 19 classes, in which 4845 images as training set and 4659 images as test set. Besides, it contains depth module information for RGB-D scene recognition.

\subsubsection{Implementation Details}
The proposed model is implemented in Pytorch \cite{paszke2019pytorch}. All models are pretrained on the Places365 \cite{zhou2017places}. The SGD optimizer is used for training in experiments. The initial learning rate is 0.01 and decreased by 10 at every 20 epochs. The momentum of the optimizer is 0.9. The total train epochs are 60. The network structure for visual scene recognition in detail is listed in Table~\ref{tab:AGCN-Visual}.

\begin{table}[]
	\centering
	\caption{Network details of AGCN for visual scene recognition. Omit the batch size.}
	\label{tab:AGCN-Visual}
	\begin{tabular}{c|c|c}
		\hline
		Layer   & Input/Input Size & Output Size  \\ \hline
		Conv1   & 3x224x224    &  64x112x112   \\ \hline
		Res-2   & 64x112x112 & 256x112x112\\ \hline
		Res-3   &   256x112x112 &  512x56x56 \\ \hline
		Res-4   &  512x56x56 & 1024x28x28\\ \hline
		Res-5   &  1024x28x28  & 2048x14x14 \\ \hline
		fc      &  2048x14x14   & 2048\\ \hline
		AFM     &  Res-4, Res-5& 1024x28x28\\ \hline
		GCN     &  AFM & \#Node*256 \\ \hline
		classification fc &  GCN, fc  & \#Class \\ \hline
		
	\end{tabular}
\end{table}

\subsubsection{Comparison with Stat-of-the-arts}
First, our method is evaluated on two datasets, Places365-7 and Places365-14, the most commonly used datasets for scene recognition. We compare our proposed model with existing state-of-the-art methods. Table~\ref{tab:place365_7_sota} demonstrates the performance of the AGCN algorithm on the Places365-7. The proposed AGCN reaches the competitive performance on both datasets without the semantic head. Instead, the BORM and OTS are semantic-based methods, i.e., these methods require the strong semantic priors obtained from the semantic segmentation network. In contrast, our AGCN has no semantic prior and additional semantic branch, and still reaches the competitive performance compared with the BORM and OTS. Besides, though AGCN is based on the ResNet50 backbone, it still outperforms ResNet50 with 9\% accuracy on the Places365-7 dataset.  Moreover, as indicated in Table~\ref{tab:places_params_comp}, the proposed AGCN only has around one-fifth of parameters and about one-third GFlops as BORM, showing the efficiency of the proposed AGCN over the semantic-based methods.

\begin{figure}[t]
	\begin{center}
		\subfigure{\includegraphics[width=\linewidth]{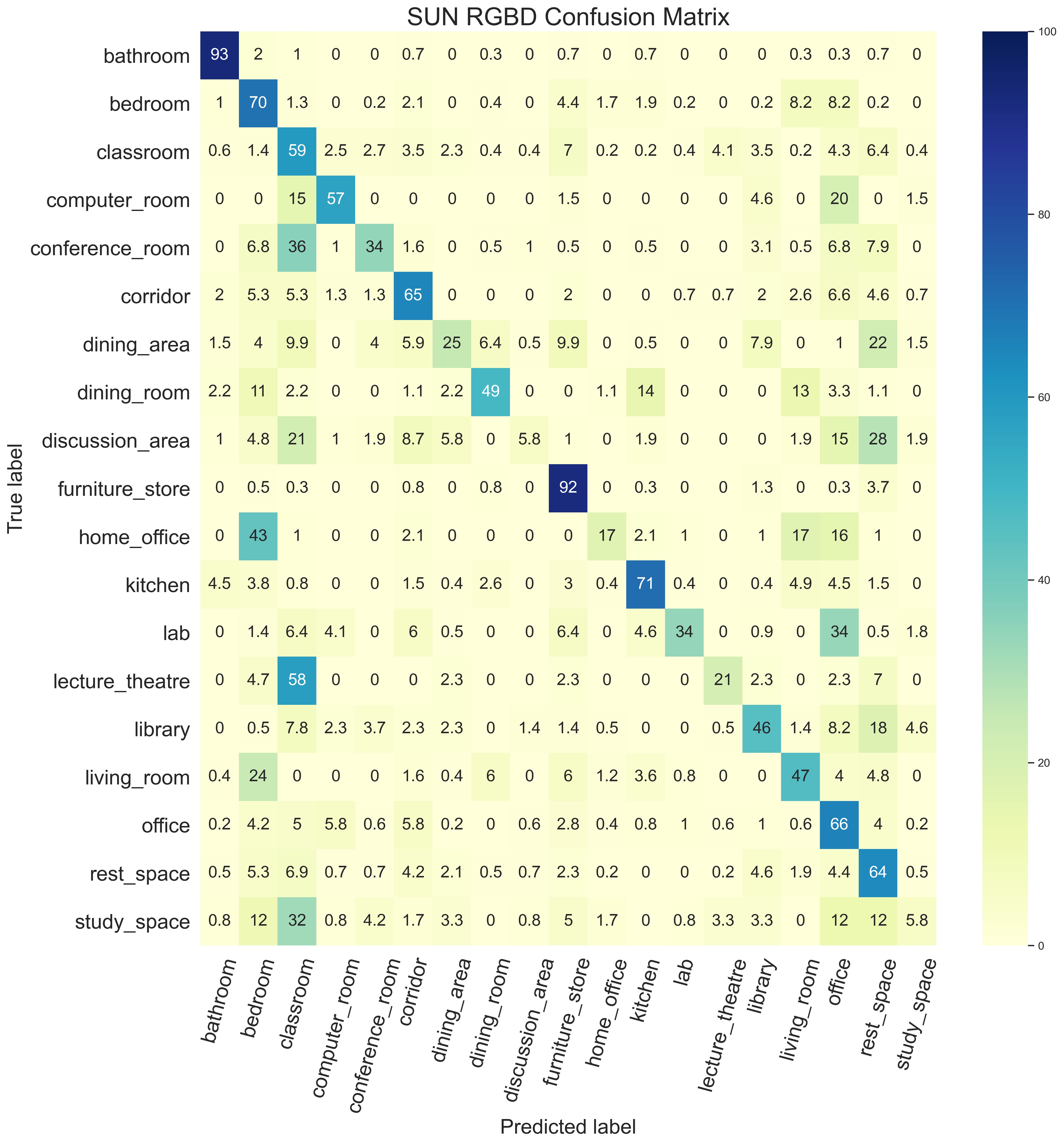}}
	\end{center}\vspace*{-10pt}\caption{Confusion Matrix for SUN RGB-D dataset. The results are shown in percentage (\%).}
	\label{fig:conf_mat_sun_rgbd}
\end{figure}

\begin{table}[htbp]
	\centering
	\scriptsize
	\centering
	\caption{Comparison with the state-of-the-art methods on the reduced Places365-7 Dataset}
	\label{tab:place365_7_sota}
	\begin{tabular}{c|ccc}
		\hline
		Method     & Config  & Semantic Head & Acc  \\ \hline
		\multirow{2}{*}{PlacesCNN \cite{zhou2017places}}     & ResNet18  & \XSolidBrush  & 80.4    \\
		& ResNet50  & \XSolidBrush  & 82.7    \\ \hline
		\multirow{3}{*}{Deduce \cite{pal2019deduce}}     &  $\Phi_{obj}$ (OM)  & \Checkmark    & 62.6       \\
		& $\Phi_{scene}$  & \XSolidBrush  & 87.3    \\
		& $\Phi_{comb.}$  & \Checkmark  & 88.1    \\ \hline
		\multirow{3}{*}{BORM \cite{zhou2021borm}} 	
		& IOM   & \Checkmark & 82.4    \\
		& BORM & \Checkmark  & 83.1    \\ 
		& CBORM  & \Checkmark & 90.1    \\	\hline
		\multirow{1}{*}{OTS \cite{miao2021object}} 	
		& OTS  & \Checkmark & 90.1    \\	\hline			
		\multirow{1}{*}{Ours} 	
		& AGCN 		 & \XSolidBrush & \textbf{91.7} 		\\ 	\hline
	\end{tabular}
\end{table}

\begin{table}[htbp]
	\hspace{1cm}
	
	\begin{minipage}[t]{.44\textwidth}
		\centering
		\caption{Comparison with the state-of-the-art methods on the reduced Places365-14 Dataset of scene recognition accuracy, the * indicates the re-implement of the method. }
		\label{tab:place365_14_sota}
		\begin{tabular}{c|ccc}
			\hline
			Method                    & Config    & Semantic Head       & Acc \\ \hline
			\multirow{2}{*}{PlacesCNN \cite{zhou2017places}}     & ResNet18 & \XSolidBrush & 76.0  \\
		& ResNet50 & \XSolidBrush & 80.0   \\ \hline
			\multirow{1}{*}{Word2Vec \cite{chen2019scene}} 
			& ResNet50+Word2Vec    & \Checkmark     & 83.7    \\ \hline
			\multirow{1}{*}{{*}Deduce \cite{pal2019deduce}} 
			
			& $\Phi_{obj}$ (OM) 	& \Checkmark    	   & 47.0	 \\ \hline
			\multirow{3}{*}{BORM \cite{zhou2021borm}}  
			& IOM       & \Checkmark      & 74.1    \\
			& BORM      & \Checkmark      & 74.9    \\ 
			& CBORM     & \Checkmark      & 85.8	\\	\hline
			\multirow{1}{*}{OTS \cite{miao2021object}} 	
			& OTS & \Checkmark   & 85.9    			\\	\hline				
			\multirow{1}{*}{Ours} 	
			& AGCN 		& \XSolidBrush & \textbf{86.0}		\\  \hline
		\end{tabular}
	\end{minipage}
\end{table}

In addition, the proposed AGCN method has been evaluated on the SUN RGB-D dataset. The confusion matrix is shown in Figure~\ref{fig:conf_mat_sun_rgbd}. It can be observed that AGCN is capable of recognizing the bathroom, bedroom, furniture store, kitchen, office, and rest space, and is not able to recognize the study space and discussion area. Moreover, home office is misclassified as bedroom, lecture theater and conference rooom are misclassified as classroom.

Table \ref{tab:sun_rgbd} shows the results on the SUN RGB-D dataset, where our method reaches competitive performance compared with other SOTA methods without additional depth information, suggesting that our method is superior to others on various scene recognition datasets. In comparison with these methods, the IOD \cite{miao2021object} utilizes the object detection results as an additional branch to boost the scene recognition performance. The MAPNet \cite{li2019mapnet}, G+L+SOOR \cite{song2019image}, TRecgNet Aug \cite{du2021cross}, and ASK \cite{xiong2021ask} are using the RGB-D information and additional branch to represent depth information, which is more complex.

\begin{table}[]
	\centering
	\caption{Comparison with the state-of-the-art methods on the SUN RGB-D dataset of recognition accuracy}
	\label{tab:sun_rgbd}
	\begin{tabular}{l|ccc}
		\hline
		Methods 				            &  RGB & Depth & RGB-D  \\ \hline
		CNN+SVM	\cite{Zhu2016}			    & 37.0 & -    & 41.5  \\
		MCAF \cite{wang2016modality}        & 40.4 & 36.3 & 48.1  \\
		MSMM \cite{song2017combining}		& 41.5 & 40.1 & 52.3  \\				         
		RGB-D-CNN \cite{song2017depth}	    & 42.7 & 42.4 & 52.4  \\
		$DF^2Net$	\cite{li2018df}		    & -    & -    & 54.6  \\
		GAN \cite{du2018depth}              & 42.6 & 53.3 & 53.3  \\
		ACM Graph \cite{yuan2019acm}        & 45.7 & -    & 55.1  \\
		G+L+SOOR \cite{song2019image}       & 50.5 & 44.1 & 55.5  \\ 
		MAPNet \cite{li2019mapnet}          & -    & -    & 56.2  \\ 
		MSN \cite{xiong2020msn}				&  -   &  -   & 56.2  \\
		ASK \cite{xiong2021ask}				& -    & - 	  & 57.3  \\
		TRecgNet Aug 	\cite{du2021cross}	& 53.8 & 49.3 & 58.5  \\
		IOD \cite{pereira2021deep}          & 58.2 & -    & 58.2  \\ \hline
		AGCN 			                    & \textbf{58.7}  & - & - \\ \hline
	\end{tabular}
\end{table}

\subsubsection{Network Performance Analysis}
From Table~\ref{tab:places365_7_gcn}, the AGCN with 20 nodes reaches best results at Places365-7 dataset. Generally speaking, increasing the number of nodes is helpful for the performance improvements of visual scene recognition.

\begin{table}[]
	\centering
	\caption{Comparison of the proposed AGCN algorithm with different number of nodes on Places365-7 dataset.}
	\label{tab:places365_7_gcn}
	\begin{tabular}{lcc}
		\hline
		Model & Places365-7                       \\  \hline
		AGCN (8 nodes)  & 91.3                     \\
		AGCN (12 nodes) & 91.4                      \\
		AGCN (16 nodes) & 91.0                      \\ 
		AGCN (20 nodes) & \textbf{91.7} 			  \\ 
		AGCN (24 nodes) & 91.3             \\ \hline
	\end{tabular}
\end{table}

As Table~\ref{tab:places_params_comp} shows, the proposed AGCN is an end-to-end training network with a much smaller number of parameters compared with the semantic segmentation-based BORM methods. The BORM, which utilizes the semantic segmentation methods, requires strong priors of the semantic information, as well as the statistical analysis of datasets for training. Our method is end-to-end training, which is more convenient for training and has fewer parameters and calculations for both network training and inference. BORM has 6x parameters and 3x GFlops compared with AGCN. In comparison with OTS, it has 3x parameters and similar GFlops. These results demonstrated our proposed methods feature fewer parameters.

\begin{figure}[t]
	\begin{center}
		\subfigure{\includegraphics[width=2.8cm]{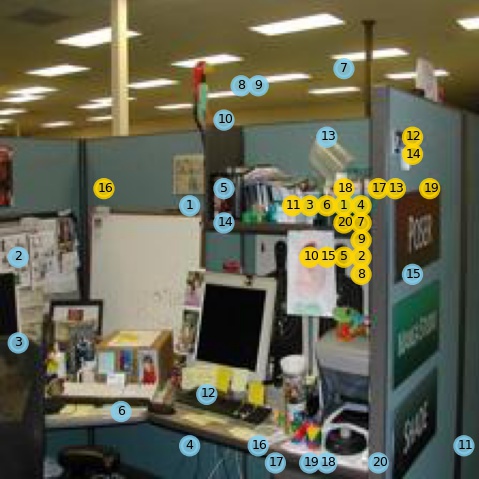}} \hspace{0pt}	% office
		\subfigure{\includegraphics[width=2.8cm]{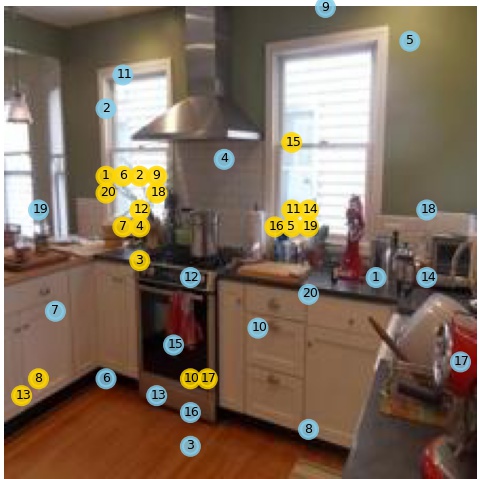}} \hspace{0pt}	% kitchen
		\subfigure{\includegraphics[width=2.8cm]{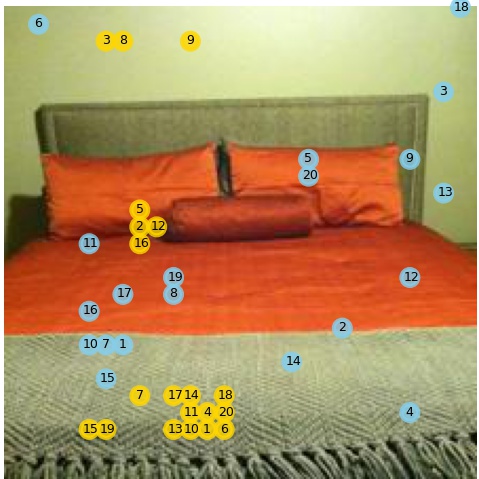}} \hspace{0pt}	% bedroom
		\subfigure{\includegraphics[width=2.8cm]{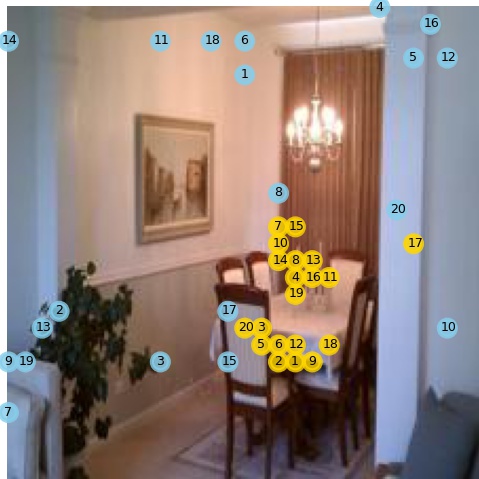}} \hspace{0pt}	% dining_room
		\subfigure{\includegraphics[width=2.8cm]{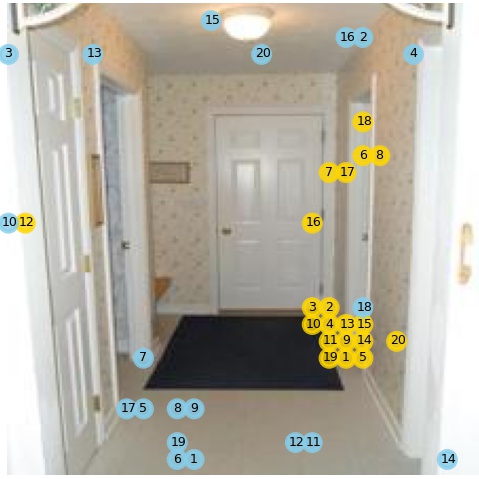}} \hspace{0pt}	% corridor
		\subfigure{\includegraphics[width=2.8cm]{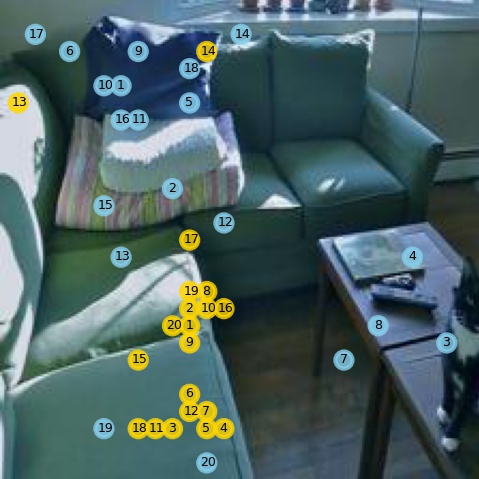}} \hspace{0pt}	% living_room
		\subfigure{\includegraphics[width=2.8cm]{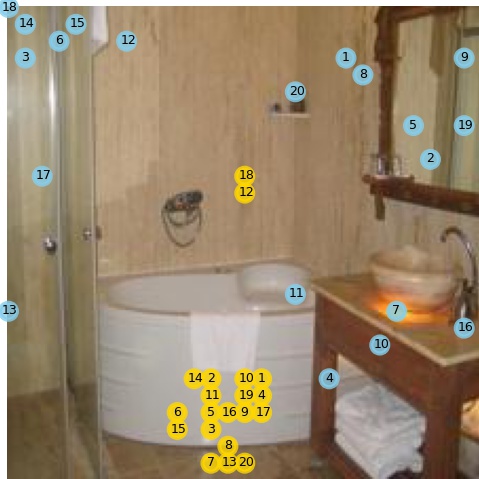}} \hspace{0pt}	% bathroom
	\end{center}\vspace*{-5pt}\caption{The scene graph visualization samples for test set of Places365-7 with 20 nodes configuration. The first row are office and kitchen, the second row are bedroom and dining room, the third row are corridor and living room. The last row is bathroom. The gold circle represent the nodes of salient acoustic graph while blue circle represent the nodes of contextual acoustic graph.}
	\label{fig:sg_visualize}
\end{figure}

\begin{figure*}[t]
	\begin{center}
		\subfigure{\includegraphics[width=3.2cm]{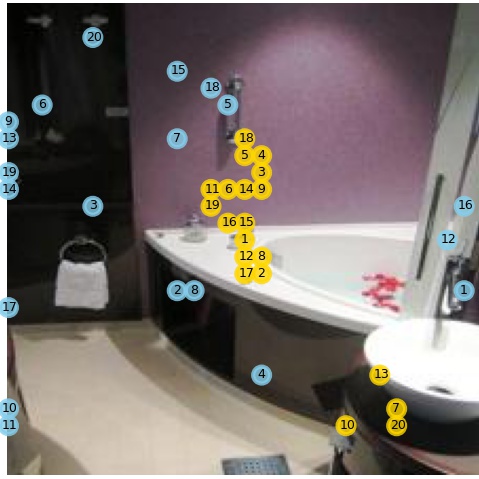}}  % Kitchen
		\subfigure{\includegraphics[width=3.2cm]{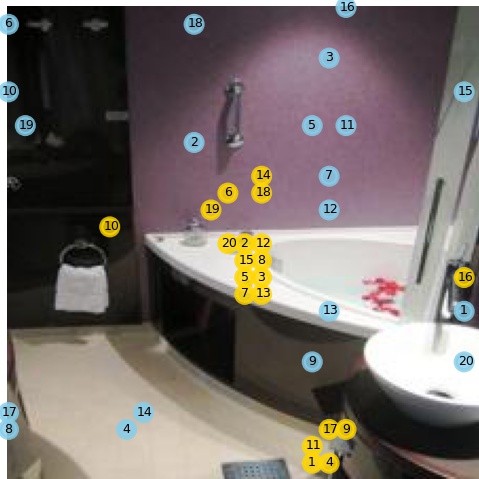}}	
		\subfigure{\includegraphics[width=3.2cm]{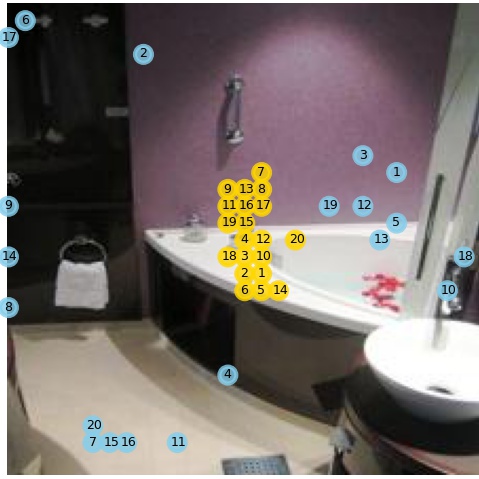}}	
		\subfigure{\includegraphics[width=3.2cm]{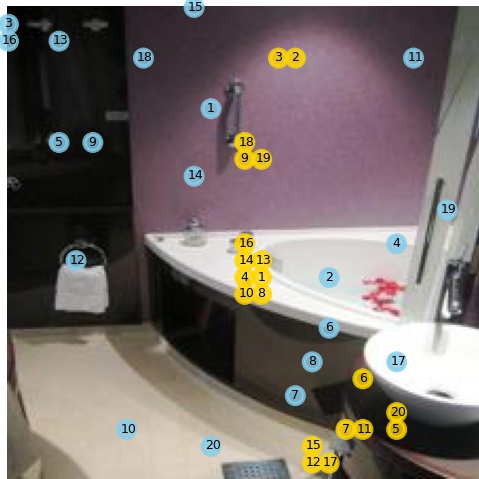}}	
		\subfigure{\includegraphics[width=3.2cm]{Figs//results_20_91_28/Places365_val_00019364.jpg}}	
		%Places365_val_00004746.jpg
		\subfigure{\includegraphics[width=3.2cm]{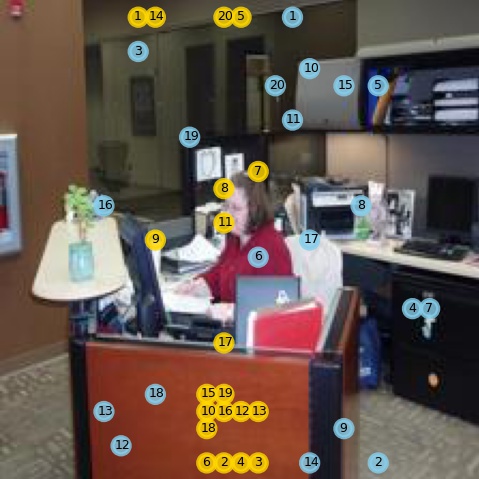}}  % office
		\subfigure{\includegraphics[width=3.2cm]{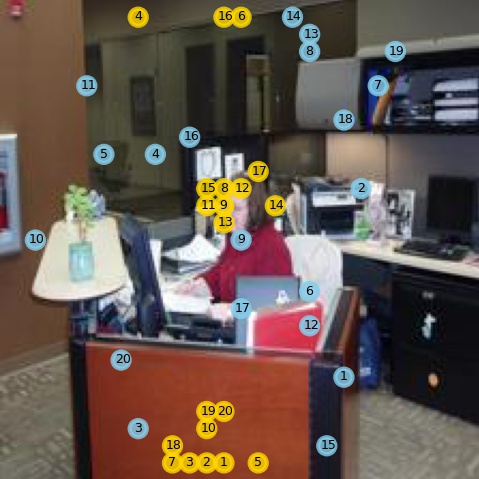}}
		\subfigure{\includegraphics[width=3.2cm]{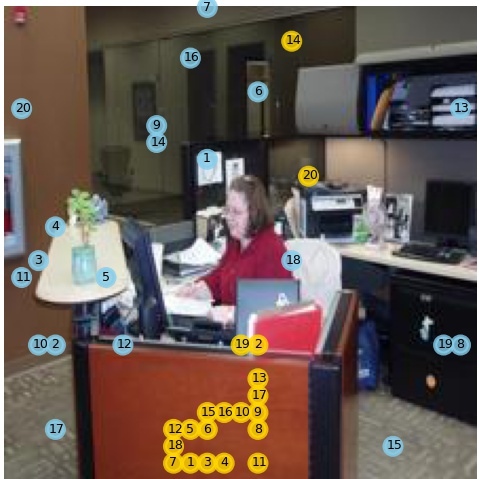}}
		\subfigure{\includegraphics[width=3.2cm]{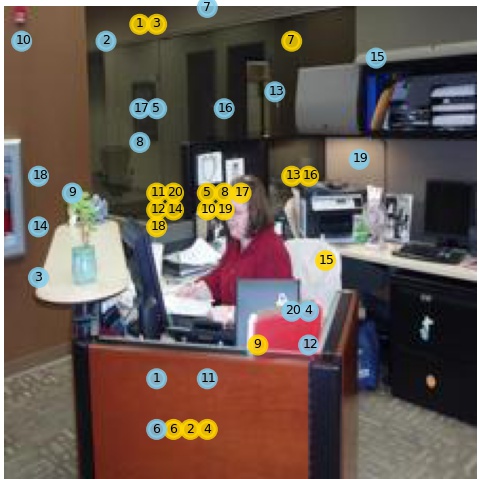}}
		\subfigure{\includegraphics[width=3.2cm]{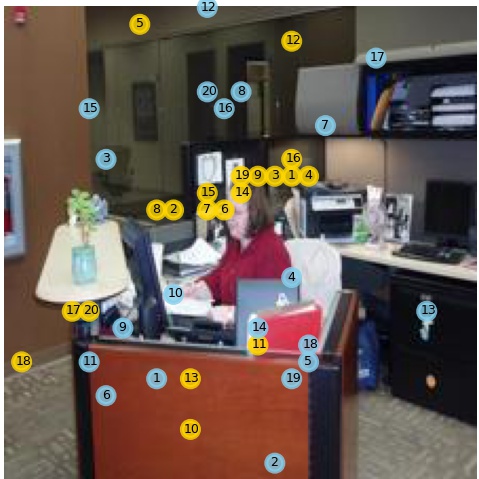}}
		
		\subfigure{\includegraphics[width=3.2cm]{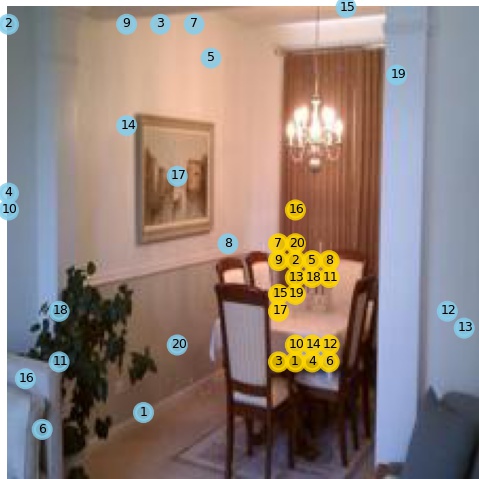}}  % dining room
		\subfigure{\includegraphics[width=3.2cm]{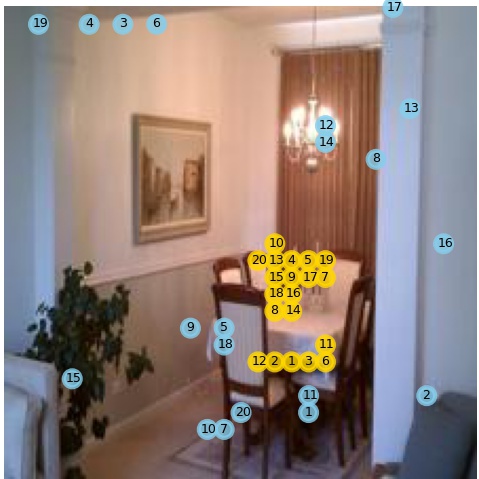}}
		\subfigure{\includegraphics[width=3.2cm]{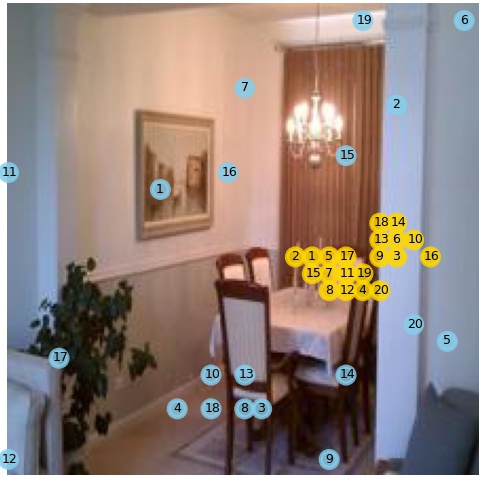}}
		\subfigure{\includegraphics[width=3.2cm]{Figs//results_20_90_57/Places365_val_00011164.jpg}}
		\subfigure{\includegraphics[width=3.2cm]{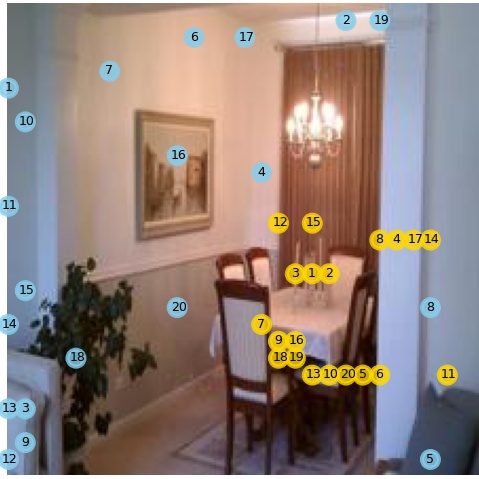}}
		
		\subfigure{\includegraphics[width=3.2cm]{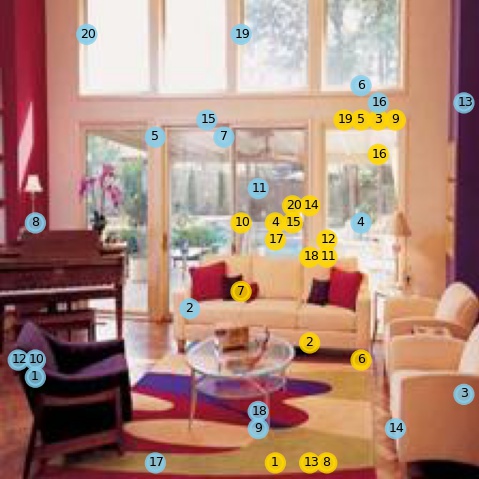}}  % living_room
		\subfigure{\includegraphics[width=3.2cm]{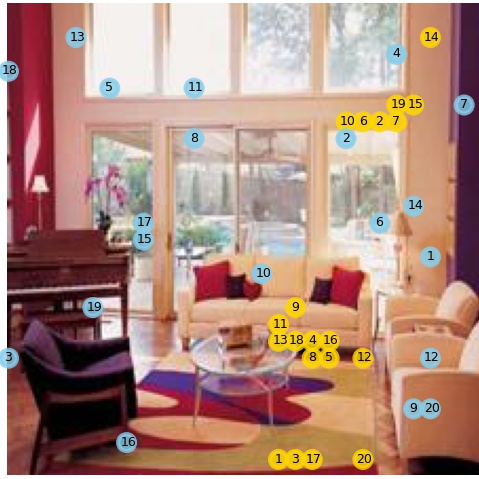}}
		\subfigure{\includegraphics[width=3.2cm]{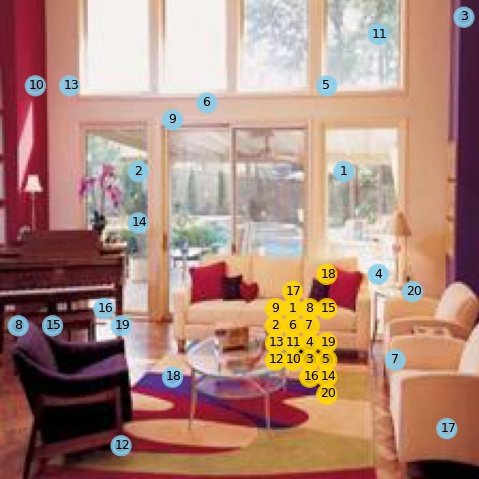}}
		\subfigure{\includegraphics[width=3.2cm]{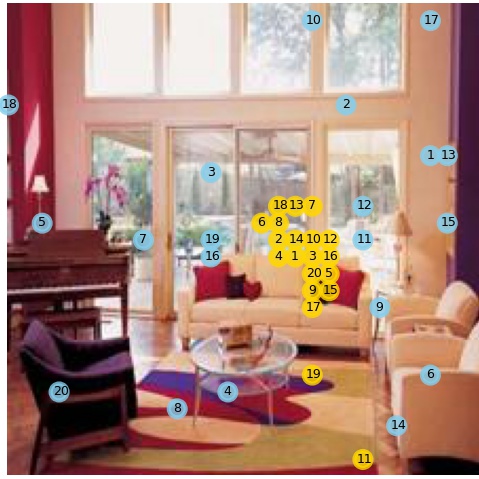}}
		\subfigure{\includegraphics[width=3.2cm]{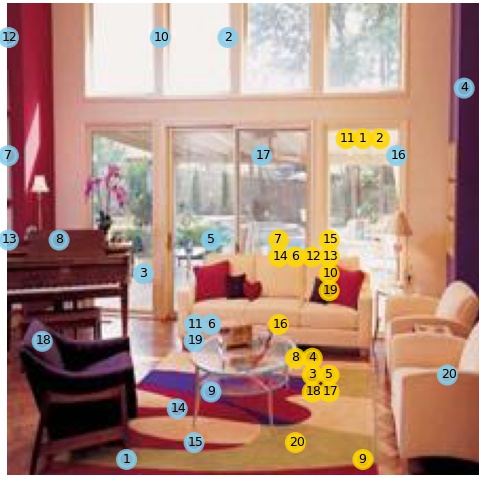}}
		
	\end{center}\vspace*{-10pt}\caption{The scene graph visualization of Places365-7 under different training epoch number. From first row to last row are bathroom, office, dining room and living room, respectively. As the training epochs increases, the salient acoustic graph is more concentrated on the important region of interests of the visual scene, while the contextual acoustic graph is scatted around the entire images for details preservation.}
	\label{fig:sg_process}
\end{figure*}

\begin{figure}[t]
	\begin{center}
		\subfigure{\includegraphics[width=3.5cm]{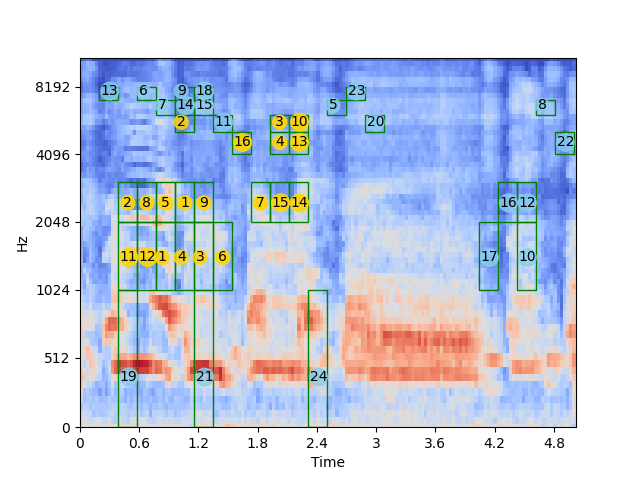}}
		\subfigure{\includegraphics[width=3.5cm]{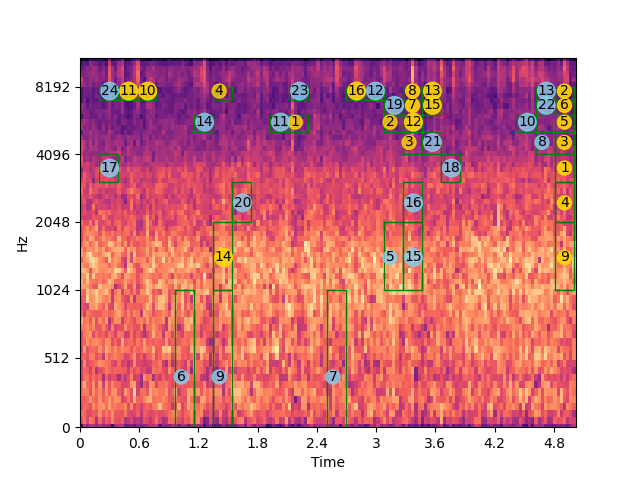}}
		\subfigure{\includegraphics[width=3.5cm]{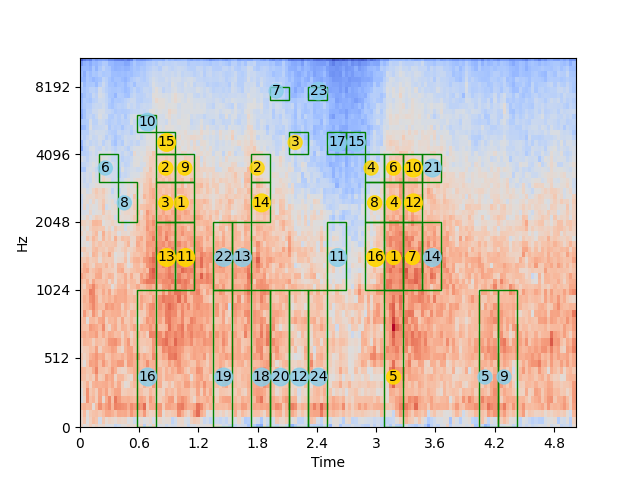}}
		\subfigure{\includegraphics[width=3.5cm]{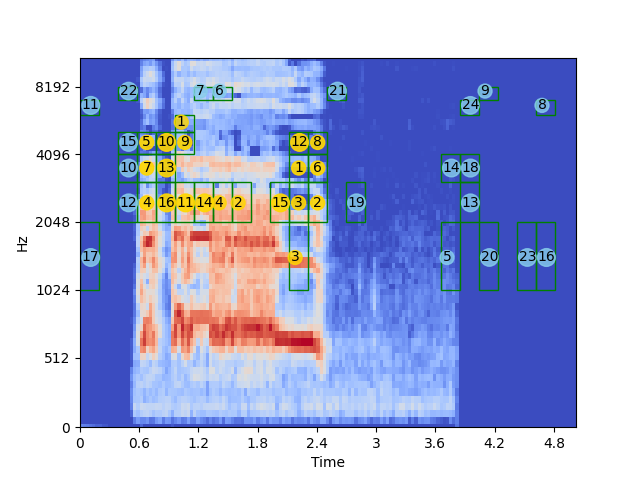}}
		\subfigure{\includegraphics[width=3.5cm]{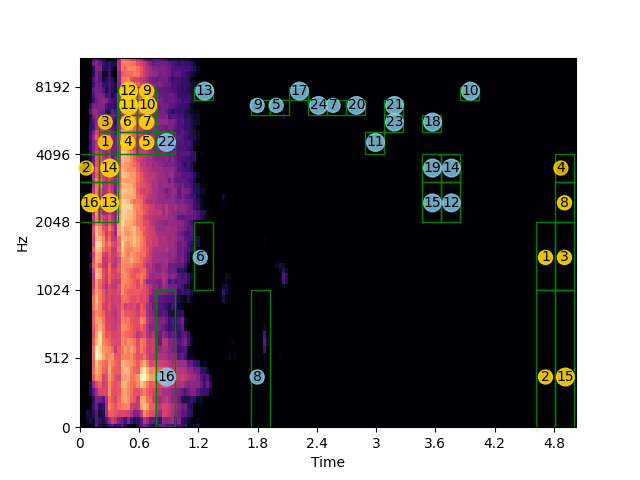}}
		\subfigure{\includegraphics[width=3.5cm]{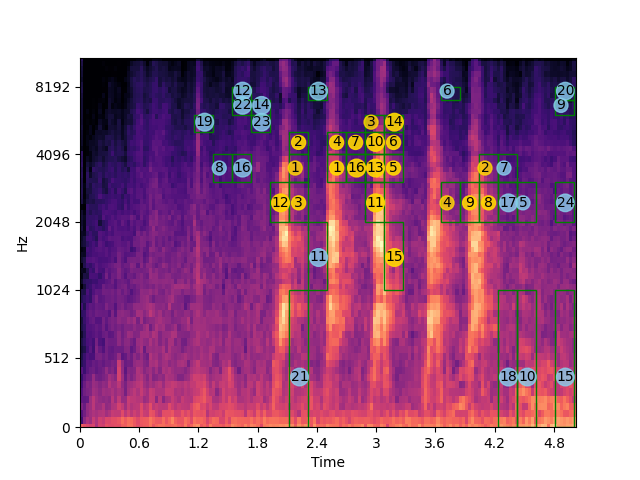}}
		\subfigure{\includegraphics[width=3.5cm]{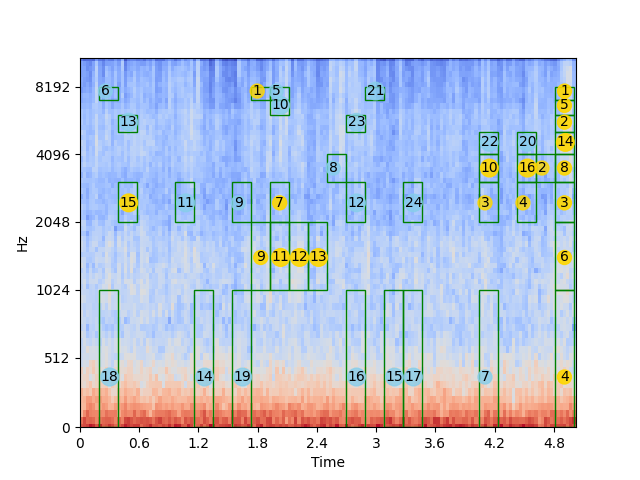}}
		\subfigure{\includegraphics[width=3.5cm]{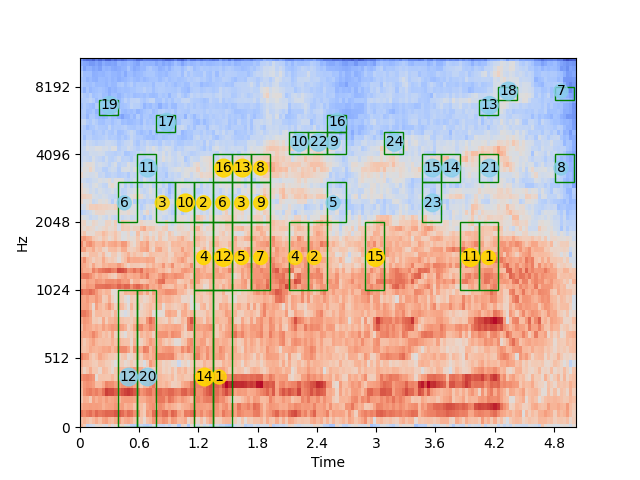}}
		\subfigure{\includegraphics[width=3.5cm]{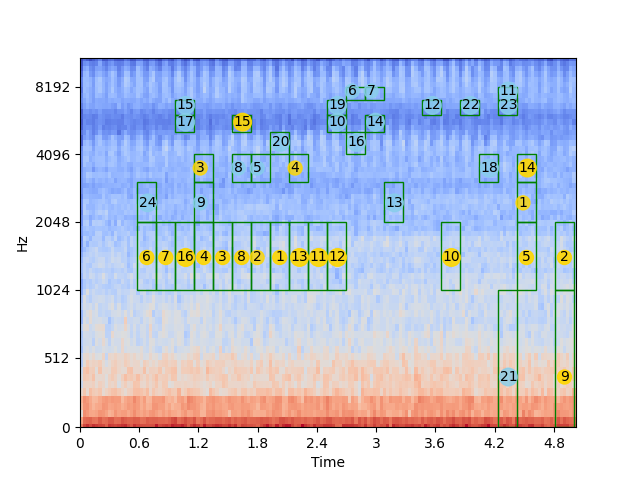}}
		\subfigure{\includegraphics[width=3.5cm]{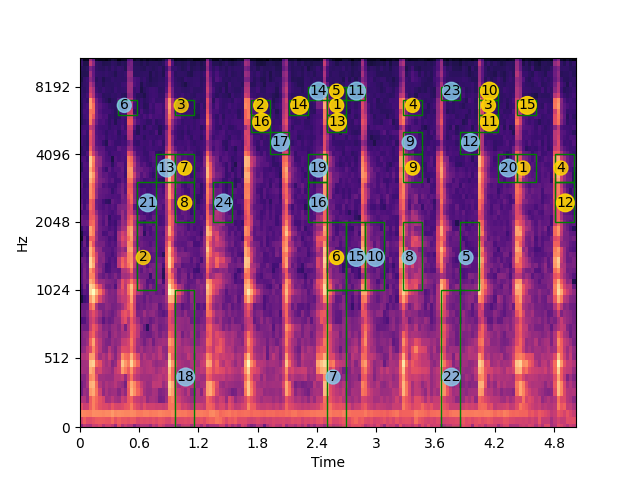}}
		
	\end{center}\vspace*{-10pt}\caption{Acoustic graphs visualization samples of ESC-10 dataset with 20 nodes configuration. From left to right column, first row to last row order, the acoustic scene classes are crying baby, rain, sea waves, roster, sneezing, dog, crackling fire, chainsaw, helicopter, and clock tick, respectively. The gold circle represents nodes of SAG, and blue circle represents the nodes of CAG.}
	\label{fig:esc_visualization}
\end{figure}

\begin{figure*}[t]
	\begin{center}
		\subfigure{\includegraphics[width=3cm]{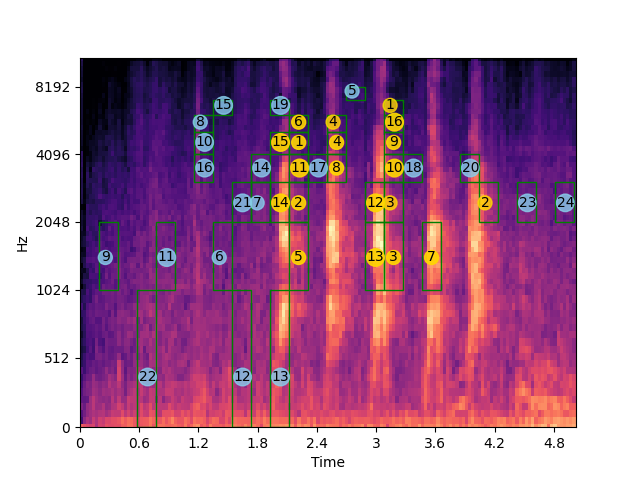}}	% 
		\subfigure{\includegraphics[width=3cm]{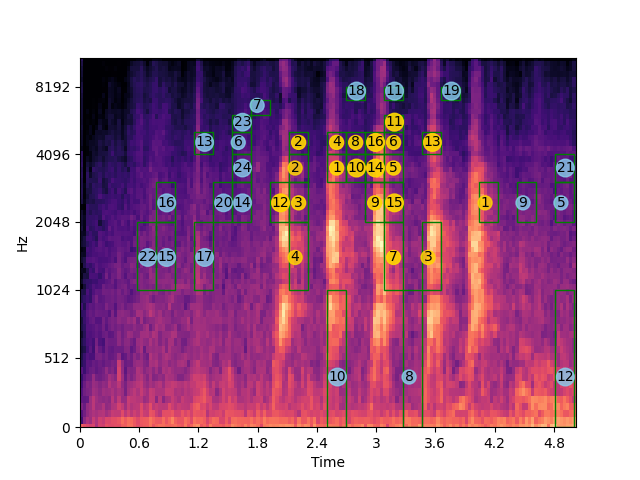}}
		\subfigure{\includegraphics[width=3cm]{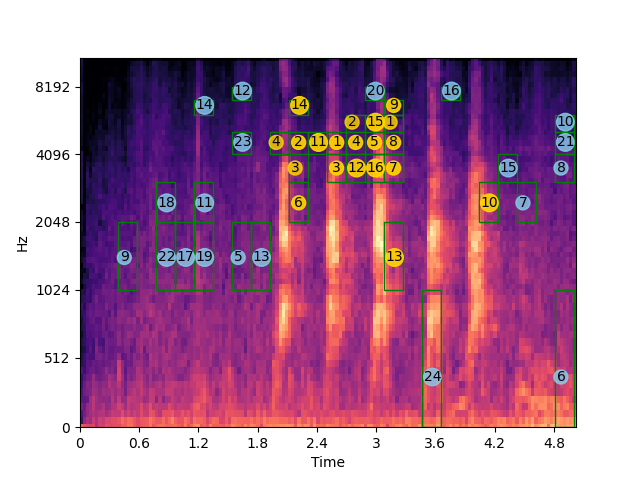}}
		\subfigure{\includegraphics[width=3cm]{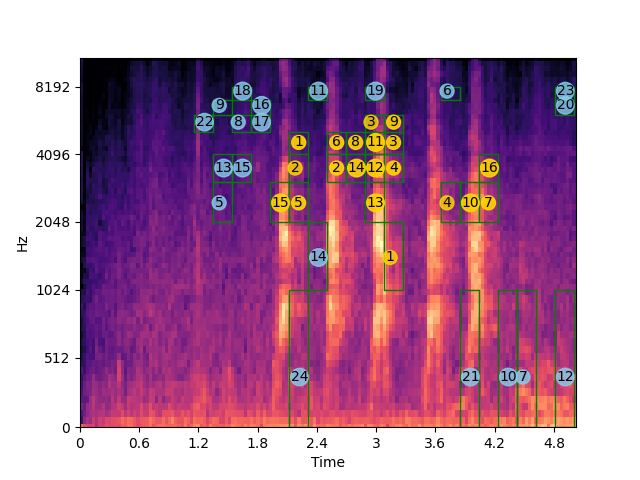}}
		\subfigure{\includegraphics[width=3cm]{Figs//epoch_44_nodes_20_acc_95_0/dog.png}}
		
		\subfigure{\includegraphics[width=3cm]{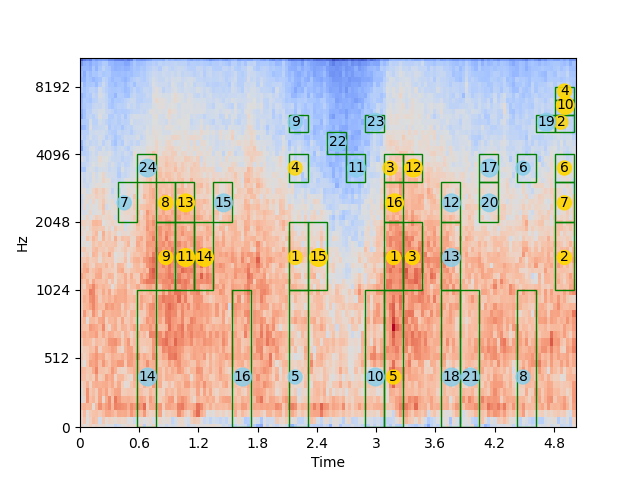}} %
		\subfigure{\includegraphics[width=3cm]{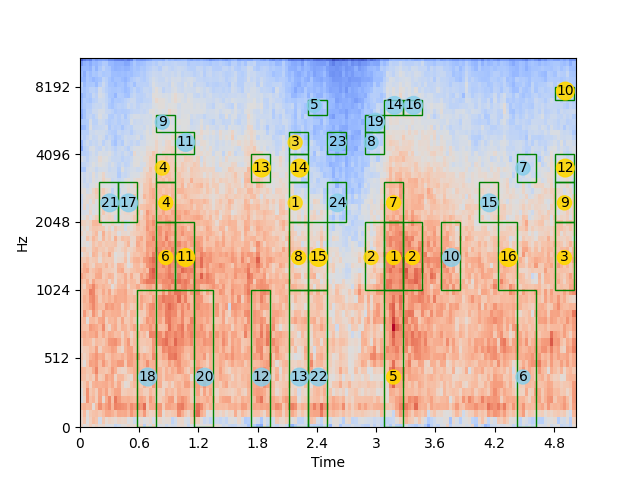}}
		\subfigure{\includegraphics[width=3cm]{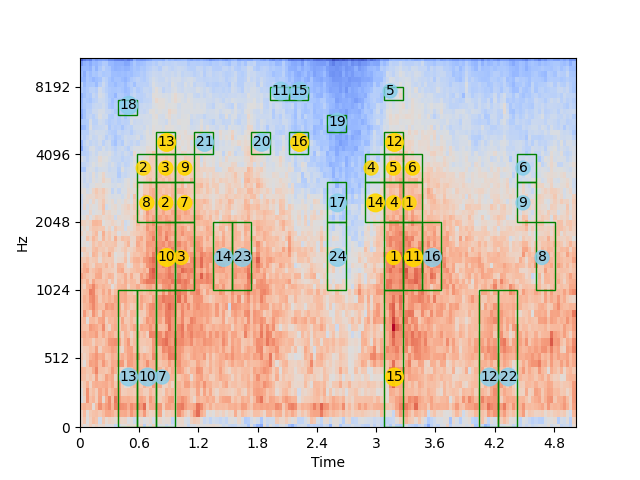}}
		\subfigure{\includegraphics[width=3cm]{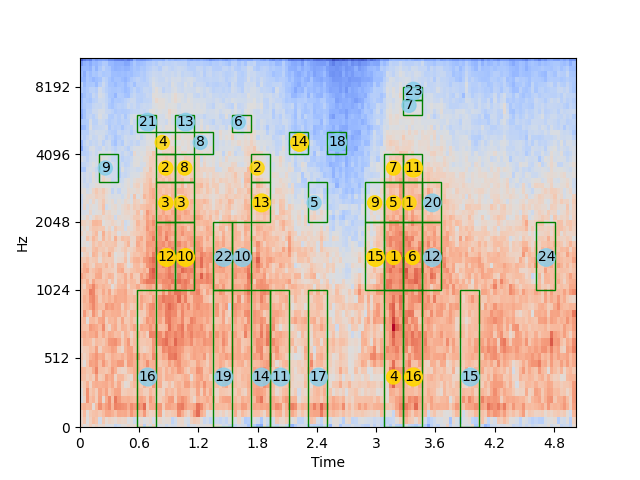}}
		\subfigure{\includegraphics[width=3cm]{Figs//epoch_44_nodes_20_acc_95_0/sea_waves.png}}

		\subfigure{\includegraphics[width=3cm]{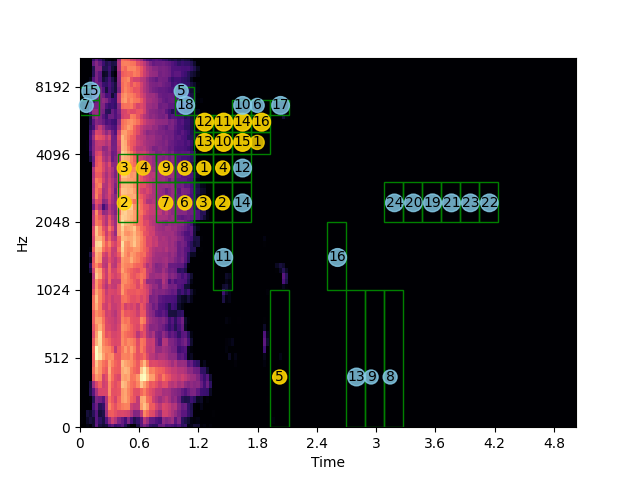}}	% 
		\subfigure{\includegraphics[width=3cm]{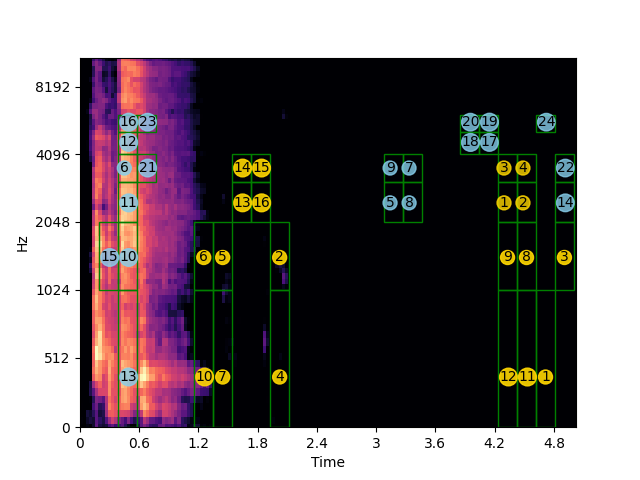}}
		\subfigure{\includegraphics[width=3cm]{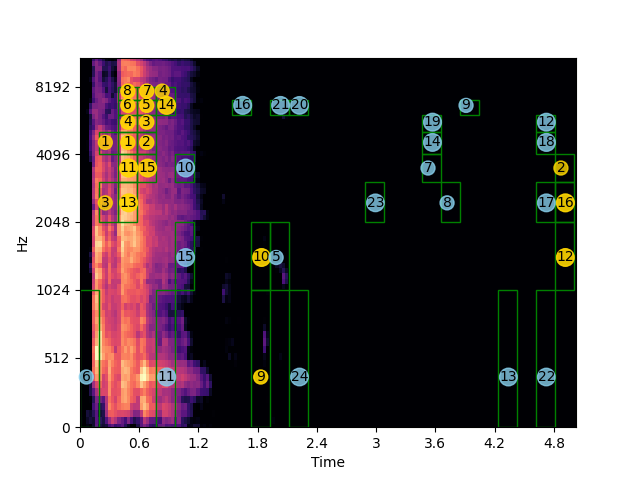}}
		\subfigure{\includegraphics[width=3cm]{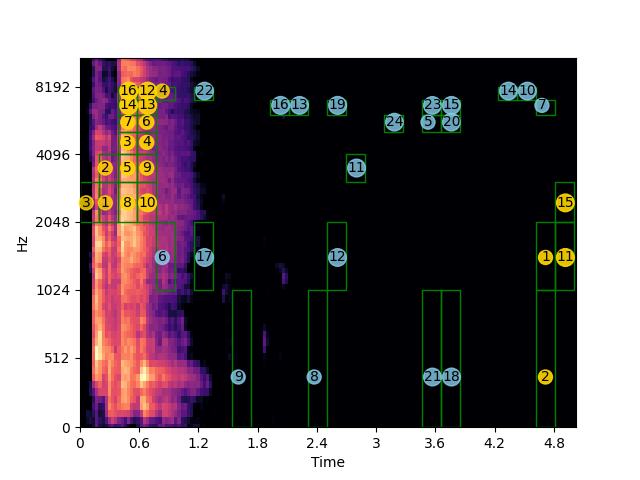}}
		\subfigure{\includegraphics[width=3cm]{Figs//epoch_44_nodes_20_acc_95_0/sneezing.png}}
		
		\subfigure{\includegraphics[width=3cm]{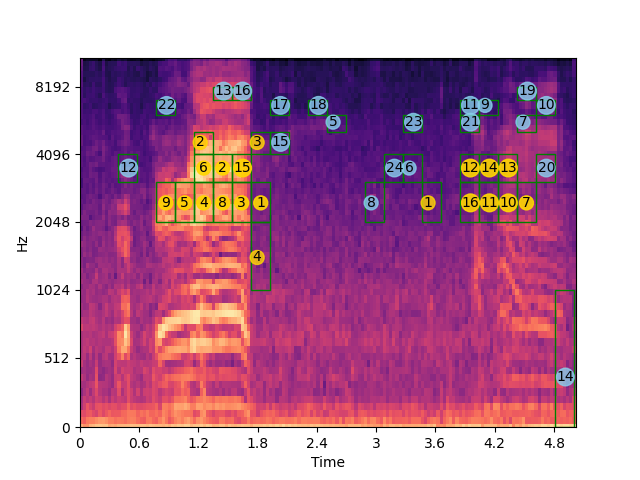}}	% 
		\subfigure{\includegraphics[width=3cm]{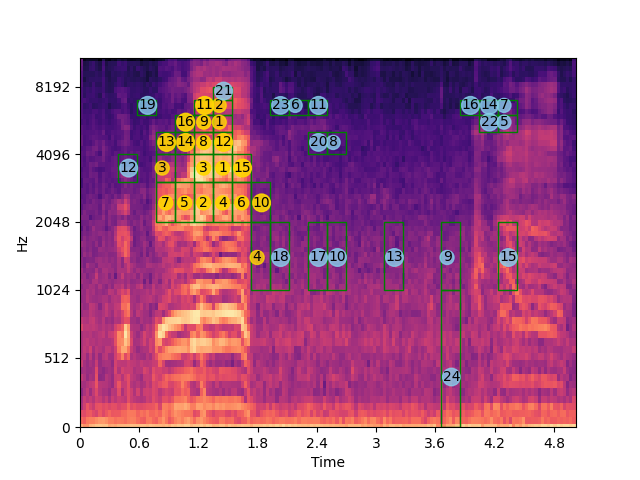}}	
		\subfigure{\includegraphics[width=3cm]{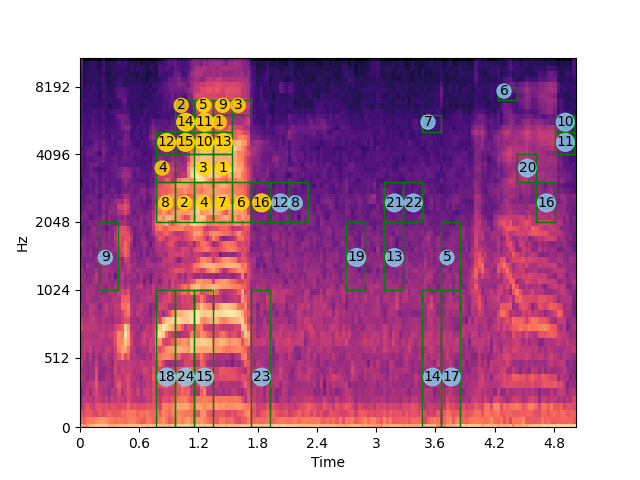}}	
		\subfigure{\includegraphics[width=3cm]{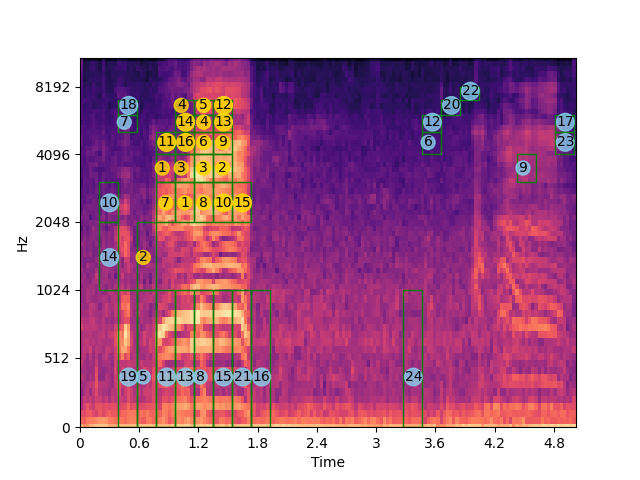}}	
		\subfigure{\includegraphics[width=3cm]{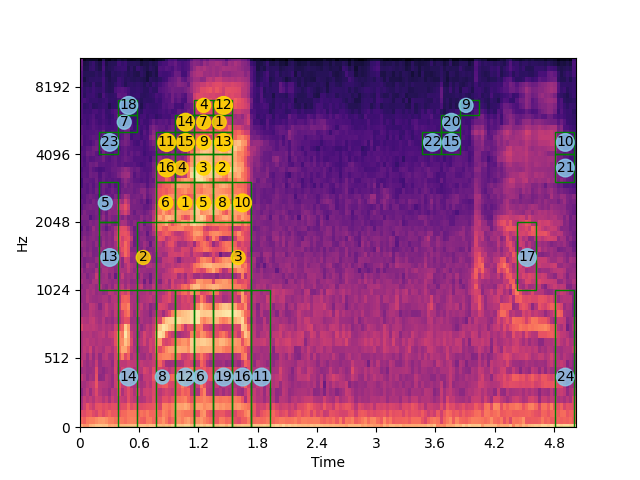}}	
	\end{center}\vspace*{-10pt}\caption{The visualization of acoustic graphs under different training epochs. The first three rows are roster, sea waves, and sneezing of ESC-10 dataset. The last row is from ESC-50, which can be viewed as unseen acoustic signal. From first row to last row are rain, roster, sea waves, and helicopter, respectively. As the training epoch increases, the salient acoustic graph is more concentrated on the important regions of the acoustic scene, while the contextual acoustic graph is evenly distributed over the entire frequency bands for details preservation.}
	\label{fig:esc_sg_process}
\end{figure*}

\begin{table}[]
	\centering
	\caption{Params Comparison}
	\label{tab:places_params_comp}
	\begin{tabular}{l|ccc}
		\hline
		Model & Params 	  & 		GFlops           \\ \hline
		BORM	\cite{zhou2021borm} &	226.89	  &   138.011        &      \\  
		Semantic Head     &    253.30 &	   46.724           			\\  
		Total 			  &    480.19 &   184.735						\\ \hline
		OTS     \cite{miao2021object}& 3.50         &       0.214          &      \\
		Semantic Head     &    253.30 &	   46.724           			\\  
		Total 			  &    256.80 &    \textbf{46.938}				\\ \hline
		AGCN (8 nodes)  &   \textbf{79.17}  &    54.729  \\
		AGCN (12 nodes) &    81.27  &    54.733  \\
		AGCN (16 nodes) &    83.37  &    54.738  \\ 
		AGCN (20 nodes) &    85.46  &    54.742  \\ 
		AGCN (24 nodes) &    87.56  &    54.746  \\ \hline
	\end{tabular}
\end{table}

\subsubsection{Visualization of Visual Scene Graph}

We show the visualization of automatically constructed SVG and CVG, as shown in Figure~\ref{fig:sg_visualize}. The visualization analysis of the graphs shows how our method focuses on the important regions and contextual information of the scene, respectively. It is worth noticing that the selected SVG concentrates more on the crucial areas, while the CVG concentrates on the contextual information, leaving out the less relevant information for scene recognition.

For example, in the second row of Figure~\ref{fig:sg_visualize}, for the left one, an image of a bedroom, the SVG is mainly focused on the bed, especially the area of sheets and pillow, while fewer nodes are on the wall. The selected CVG is scattered around the entire image to evenly capture more detailed information on sheets, pillows, and walls. Similarly, for the right one, a picture of the Kitchen, the selected SVG is focused more on the dining table area while paying less attention to the wall and floor. At the same time, the CVG is located around each area evenly for detail preservation. 

We have visualized the scene graph during the training process. It is worth mentioning that as the network training converges, a better distribution of scene graphs is obtained. The scene graph is better at mimicking the spatial layout of the scene objects or important regions of the scene. The visualization of the training process is shown in Figure~\ref{fig:sg_process}. The first row shows the visualization of a bathroom with training epochs increases. It is worth noticing that the SVG focuses on several important elements of a bedroom like a bathtub, washbasin, walls, and floors instead of solely focusing on the bathtub. The second row shows the visualization of an office. In the beginning, the SVG is located in less essential regions like the area of a wooden panel of a desk. Later on, the SVG is moved to the working area near the printing machine, computer screen, and office chair, showing the excellent adaptivity of the proposed algorithm during training. The third row is a dining room, where the SVG focuses on the region of the dining table, and the CVG is scatted around the dining room to capture more detailed surrounding information of the dining room as additional cues. The last row is a living room. At first, the SVG is mainly located around the big glass area and floor and gradually moves the attention to the sofa, coffee table, and floor, showing the superior learning ability to find essential regions of a scene.

\subsection{Acoustic Scene Classification}
\subsubsection{Datasets}
We use Environmental Sound Classification (ESC) dataset that contains 50 acoustic scene classes for evaluation \cite{piczak2015esc}. The official dataset splition settings ESC-10 and ESC-50 are utilized, where the 5-fold cross validation is applied. The ESC-50 dataset contains 2000 labeled environmental sounds with a duration of 5 seconds, with each categories are equally distributed. The ESC-50 is split as training set with 1600 samples and test set with 400 samples. The ESC-10 is a subsect of ESC-50 that contains 320 samples as training set, and 80 samples as test set.

\subsubsection{Implementation Details}
The raw videos are resampled to 16.0kHz and fixed to the certain length, 5s for ESC-10 and ESC-50. Then, to analyse the signal in frequency domain, the short time Fourier transform (STFT) is applied on the audio signals, where the spectrograms is obtained with a hop length of 400 and window length of 1024. Last, to obtain the Log-Mel spectrograms, the 64 mel filter banks are applied on the previous spectrograms and followed by a logarithmic operation. During the training stage, we adopts the ResNet50 pretrained on the AudioSet for sound information extraction \cite{wang2018polyphonic}. The proposed model is implemented in Pytorch \cite{paszke2019pytorch}. The SGD optimizer is used for training in experiments. The initial learning rate is 0.01 and decreased by 10 at every 20 epochs. The momentum of optimizer is 0.9. The total train epochs is 60. The network structure in details are listed on the Table~\ref{tab:AGCN-Acoustic}. The input feature of sound signal after Fourier transform is of shape 1x201x64. Therefore, we adjust the input channel of ResNet50 from 3 to 1.

\begin{table}[]
	\centering
	\caption{Network details of AGCN for acoustic scene recognition.  Omit the batch size.}
	\label{tab:AGCN-Acoustic}
	\begin{tabular}{c|c|c}
		\hline
		Layer   & Input/Input Size & Output Size \\ \hline
		conv1   &  1x201x64    &  64x101x32         \\ \hline
		Res-2   &  64x101x32  &  256x101x32           \\ \hline
		Res-3   &  256x101x32   & 512x51x16 \\ \hline
		Res-4   &  512x51x16  & 1024x26x8  \\ \hline
		Res-5   &  1024x26x8  & 2048x13x4  \\ \hline
		fc      &    2048x13x4 & 2048  \\ \hline
		AFM     &  Res-4, Res-5   & 1024x26x8  \\ \hline
		GCN 	&  AFM & \#Node*256 \\ \hline
		classification fc & GCN, fc& \#Class\ \\ \hline
	\end{tabular}
\end{table}

\begin{figure}[t]
	\begin{center}
		\subfigure{\includegraphics[width=8cm,height=6cm]{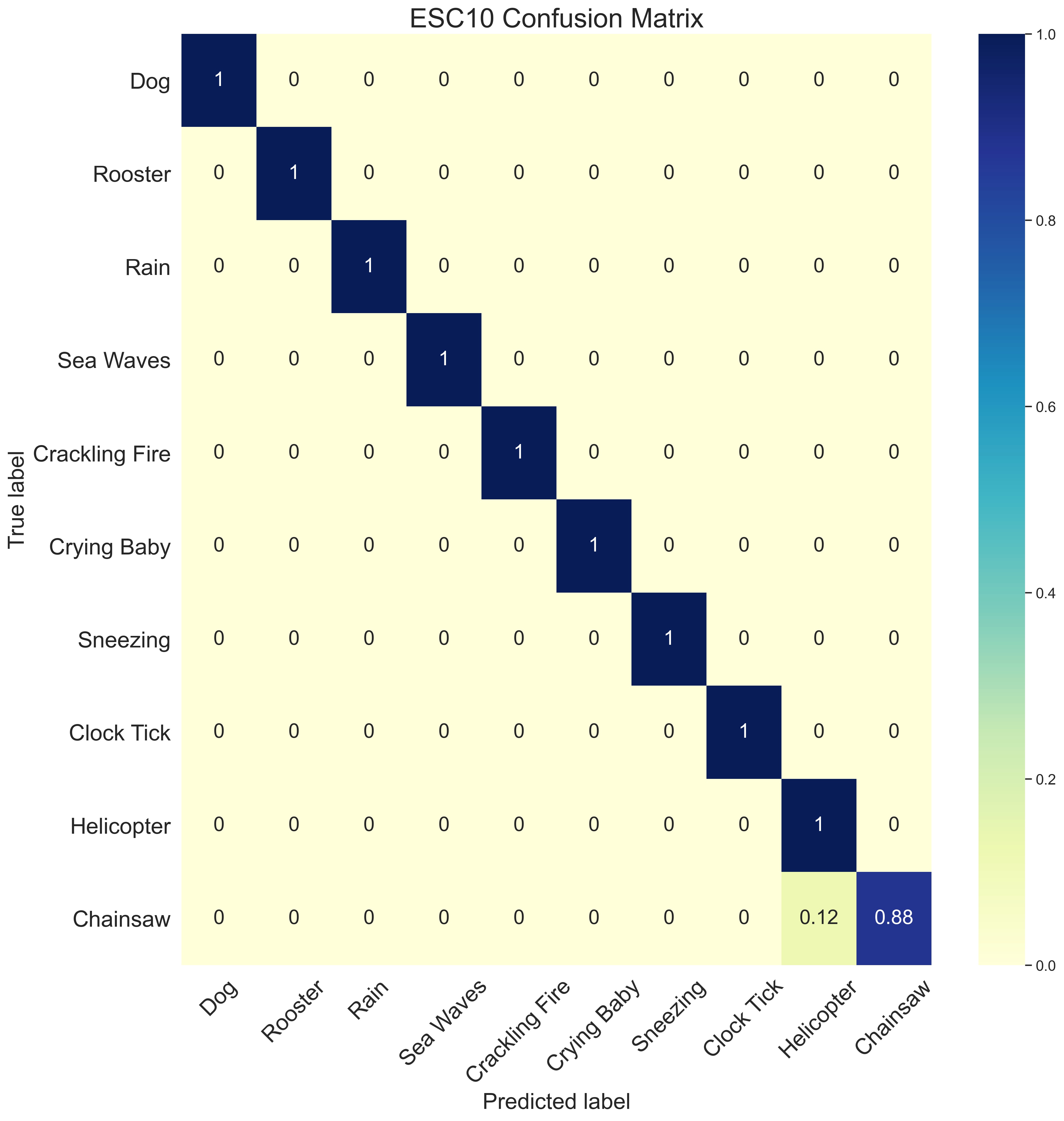}}
	\end{center}\vspace*{-10pt}\caption{Confusion Matrix for ESC-10 dataset. The results are shown in percentage (\%/100).}
	\label{fig:conf_mat_esc10}
\end{figure}

\begin{figure}[ht]
	\begin{center}
		\subfigure{\includegraphics[width=\linewidth]{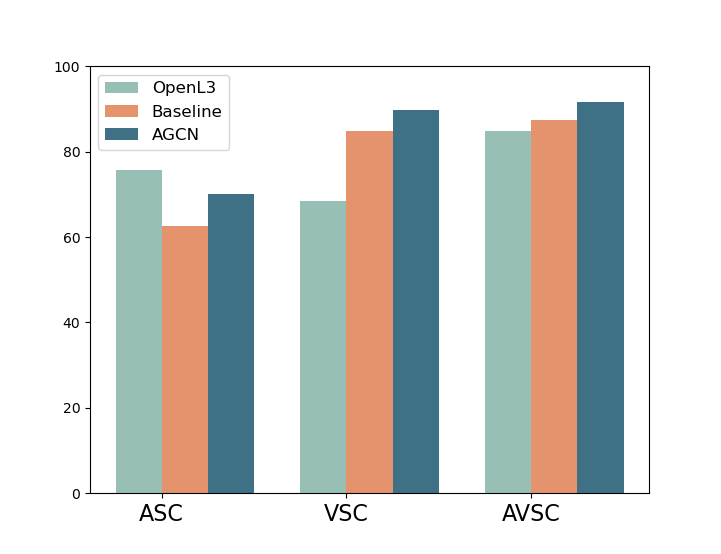}}
	\end{center}\vspace*{-10pt}\caption{The performance of OpenL3 \cite{wang2021curated} based method and AGCN on the ASC, VSC, and AVSC tasks.}
	\label{fig:avsc}
\end{figure}

\begin{figure*}[!ht]
	\begin{center}
		\subfigure{\includegraphics[width=\linewidth]{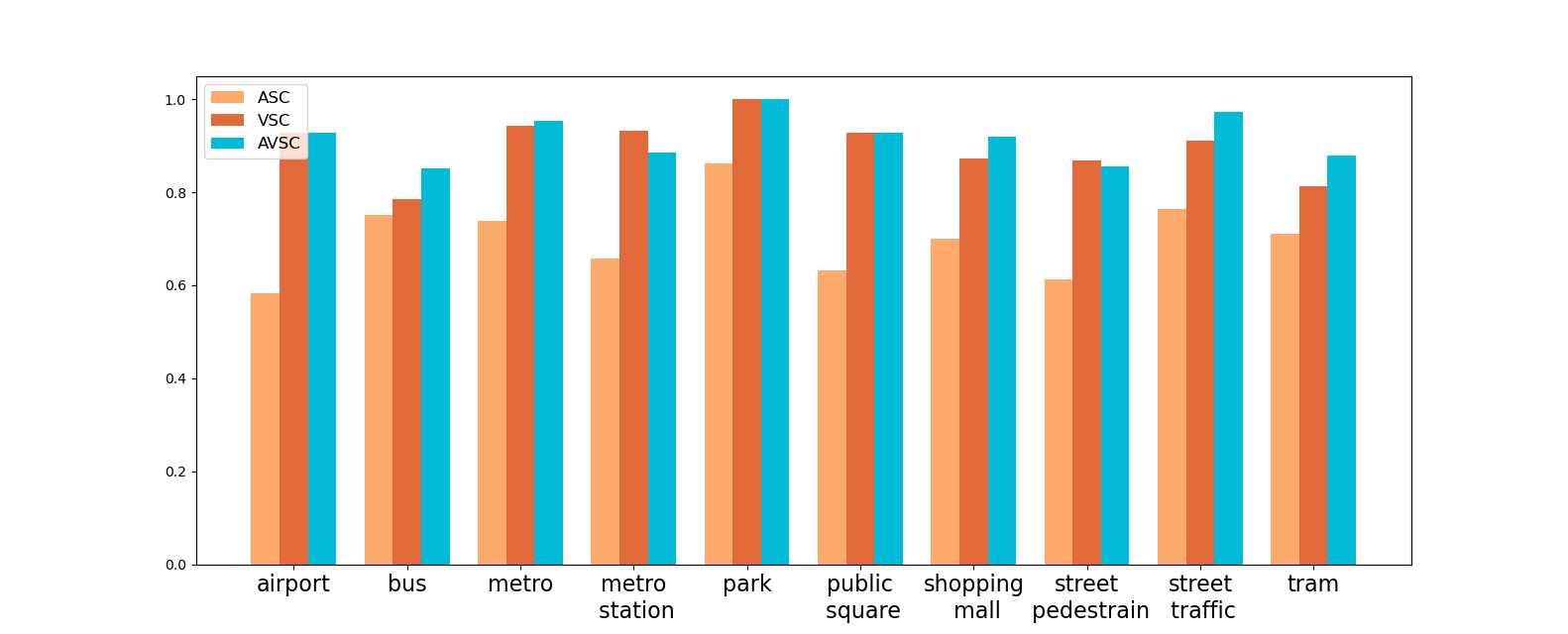}}
	\end{center}\vspace*{-10pt}\caption{The per-class accuracy performances of AGCN on the TAU Audio-Visual Scenes 2021 dataset, the ASC, VSC, and AVSC tasks are reported. }
	\label{fig:avsc_per}
\end{figure*}

\subsubsection{Comparison with stat-of-the-arts}
\begin{table}[]
	\centering
	\caption{Comparison of the AGCN and existing state-of-the-art methods on the ESC-10, ESC-50 datasets. We perform 5-fold cross-validation followed by official fold settings. The average accuracy of 5-fold cross-validation are reported.}
	\label{tab:esc}
	\begin{tabular}{lcc}
		\hline
		Model & ESC-10 & ESC-50 \\ \hline
		KNN \cite{piczak2015esc} & 66.7 & 32.3 \\ 
		SVM \cite{piczak2015esc} & 67.5 & 39.6 \\ 
		Random Forest \cite{piczak2015esc} & 72.7 & 44.3 \\ 
		AlexNet \cite{boddapati2017classifying} & 78.4 & 78.7 \\ 
		Google Net \cite{boddapati2017classifying} & 63.2 & 67.8 \\ 
		PiczakCNN \cite{piczak2015esc} & 80.5 & 64.9 \\ 
		SoundNet \cite{aytar2016soundnet} & 93.7 & 79.1 \\ 
		Multi-Stream CNN \cite{Li2019} & 94.2 & 84.0 \\ 
		MelFB+LGTFB+EN-CNN \cite{park2020cnn} & 93.7 & 88.1 \\
		Attention Network\cite{Li2019} & 94.2 & 84.0 \\
		Human\cite{piczak2015environmental} & 95.7 & 81.3 \\
		TS-CNN10 \cite{Wang2020} & 95.8 & 88.6 \\ 
		ARNN \cite{tripathi2021environment} & 92.0 & - \\
		DA-KL \cite{tripathi2022data} & 92.5 & - \\
		ACRNN \cite{zhang2021attention} & 93.7 & 86.1 \\ \hline
		AGCN (Ours) & \textbf{98.0}  & \textbf{90.8} \\ \hline
	\end{tabular}
\end{table}

\begin{table}[]
	\centering
	\caption{Comparison of the proposed AGCN with different number of nodes}
	\label{tab:esc_agcn}
	\begin{tabular}{lccc}
		\hline
		Model & \# Nodes & ESC-10 & ESC-50 \\ \hline       
		AGCN & 8  & 96.3 & 88.3          \\
		AGCN & 12 & 97.5 & 90.5          \\
		AGCN & 16 & 97.5 & 89.6          \\
		AGCN & 20 & 97.0 & \textbf{91.2}          \\
		AGCN & 24 & \textbf{98.5} & {90.8} \\  \hline
	\end{tabular}
\end{table}

Our method is evaluated on two ESC datasets, ESC-10 and ESC-50, which are the most commonly used datasets for ESC, where the Log-Mel spectrograms are extracted from the audio signals as the input of the AGCN.
We compare our proposed model with existing state-of-the-art methods. Table~\ref{tab:esc} demonstrates the performance of AGCN on the ESC-10 and ESC-50 datasets. The proposed AGCN reaches state-of-the-art performance on both datasets. In the ESC-10 dataset, our method reaches 98.5\% classification accuracy, which is higher than TS-CNN10 about 2.7\% accuracy. The detailed confusion matrix of ESC-10 results are shown in Figure~\ref{fig:conf_mat_esc10}. In the ESC-50 dataset, our method reaches 90.8\% classification accuracy, which is 2.2\% higher than TS-CNN10.

\subsubsection{Effective of Node Number}
From the Table~\ref{tab:esc_agcn}, we observed that the influence of nodes numbers to the recognition accuracy is near linearly correlated. The best accuracy is with the 24 number of nodes, showing the \textbf{98.0\%} and \textbf{90.8\%} classification accuracy on the ESC-10, and ESC-50 datasets, respectively.

\subsubsection{Visualization of Acoustic Graphs}

We show the visualization of automatically constructed SAG and CAG, as shown in Figure~\ref{fig:esc_visualization}. The visualization analysis of SAG shows how our method focuses on the crucial regions of sound signal spectrograms. Besides, it shows CAG  captures the detailed patterns of sound signal spectrograms. AGCN well exploited the sound signal textures and frequency bands textures and abandoned the less relevant information for acoustic representation. For instance, in the third row, the left column shows the spectrograms of a sneezing sound spectrogram, where the SAG focuses on the essential region with a relatively strong signal, and the CAG is distributed over the entire signal frequency bands to preserve the details of acoustic signal information. The right column shows a sound signal spectrogram of the dog. It can be seen that the SAG makes use of the most discriminative signal bands while the CAG utilizes the signals of different frequencies, and strengths for a details representation of the acoustic scene.

We have visualized the acoustic graphs during the training process. It is obvious that as the network training converges, optimized distribution of acoustic graphs is obtained. The visualization of the training process is shown in Figure~\ref{fig:esc_sg_process}. The first three rows are samples of the ESC-10 dataset. The first row shows an auditory scene of the dog. In the first column, the nodes of the SAG are diversified between the contextual regions and salient regions of signal, then rapidly converge to the salient regions at the following epochs. The CAG is scatted around the entire sound signal but then filters out a small portion of the sound signal between 0-1 second, only to preserve the more useful information. The second row shows the acoustic scene of sea waves. At the early stage, like the first two columns, the SAG is diverged and distributed around the entire spectrograms and then gradually concerted on two regions, showing a convergence ability. The CAG is scatted around the frequency bands to capture the details of each frequency level, ensuring the signal is well exploited. The third row shows the sneezing scene. The SAG moves from one place to another in the first two columns, gradually focusing on the most discriminative signal regions and forming a distinctive SAG. The last row is the sheep scene from the ESC-50 dataset, viewed as an unseen acoustic scene. In the beginning, there are two locations for the SAG, which are not representative. Then, the SAG is gradually converging to the most apparent regions. The CAG is distributed on the high-frequency bands at the beginning and then spread over the entire frequency bands for detail preservation.

\subsection{Audio-Visual Scene Classification}
\subsubsection{Datasets}
TAU Audio-Visual Scenes 2021 dataset \cite{wang2021curated} is used to test the effectiveness of the proposed AGCN. Both the audio and video samples have a duration of 10 seconds . Based on the official settings, there are 86K samples in the training set, and 36K samples in the test set. The audio processing is the same as the previous ASC task. To ensure the efficiency of AGCN, we only sample one image from each video as the input for visual modalities. 

\subsubsection{Results}
Figure~\ref{fig:avsc} shows the comparison between the baseline and OpenL3-based methods, which shows a substantial improvement has been achieved. Compared with the OpenL3-based methods \cite{wang2021curated}, we achieve a relative improvement in VSC and AVSC tasks.  We have achieved 89.5\% and 91.6\% accuracy on the VSC and AVSC tasks. The relative improvement of our proposed AGCN over the OpenL3-based method on the VSC task is about  31.1\%, which is tremendously surprising. Overall, on the AVSC task, our relative improvement over the OpenL3-based method is about 8.0\%, which suggests a substantial improvement is achieved by the AGCN. Figure~\ref{fig:avsc_per} shows the per-class accuracy of AGCN on three tasks, compared with VSC based task, AVSC has better performance. These results suggest proposed AGCN is suitable for the AVSC task.

%%%%%%%%%%%%%%%%%%%%%%%%%%%%%%%%%%%%%%%%%%%%%%%%%%%%%%%%%%%%%%%%%%%%%%%%%%%%%%%%%%%%%%%%%%%%%%%%%%%%%%

\section{Conclusion}
\label{sec:conclusion}
In this paper, we propose an AGCN network for audio-visual scene representation that exploits the SAG/SVG and CAG/CVG for audio-visual scene classification. We explicitly construct the SAG/SVG that focuses on essential regions of audio-visual scene inputs with the most distinctive nodes selected from feature maps. Meanwhile, to utilize the meaningful background contextual information for better audio-visual signal modeling, we construct CAG/CVG with average nodes selected from fused feature attention maps. In addition, the spatial layouts of scene nodes are preserved with an adjacency matrix. Extensive experiments have been conducted on ASC, VSC, and AVSC tasks. These results showed the superiority of AGCN over the previous CNN-based methods. Moreover, the visualized graphs show the proposed SAG/SVG and CAG/CVG are semantic meaningful.

\ifCLASSOPTIONcaptionsoff
  \newpage
\fi



%\bibliographystyle{IEEEtran}
%\begin{thebibliography}{1}

%\newpage
\bibliographystyle{IEEEtran}
\bibliography{scene_recognition}

\end{document}